%% file: colmlm_iclr_2027.tex
\documentclass{article} %

\usepackage{iclr2027_conference,times}

\usepackage[utf8]{inputenc} %
\usepackage[T1]{fontenc}    %
\usepackage{hyperref}       %
\usepackage{url}            %
\usepackage{booktabs}       %
\usepackage{amsfonts}       %
\usepackage{amssymb}
\usepackage{amsmath}        %
\usepackage{nicefrac}       %
\usepackage{microtype}      %
\usepackage{xcolor}         %
\usepackage{graphicx}
\usepackage{subcaption}
\usepackage{multirow}
\usepackage{float}
\usepackage{wrapfig}
\usepackage{placeins}       %
\usepackage{todonotes}
\usepackage{enumitem}
\usepackage{colortbl}
\usepackage{threeparttable}
\usepackage{afterpage}
\usepackage{makecell}

\usepackage{pifont}
\usepackage{xspace}

\usepackage{xcolor}
\usepackage{tcolorbox}
\tcbuselibrary{breakable, listings}

\definecolor{darkblue}{rgb}{0, 0, 0.5}
\hypersetup{colorlinks=true, citecolor=darkblue, linkcolor=darkblue, urlcolor=darkblue}

\definecolor{clGray}{HTML}{959595}
\definecolor{clGrayBg}{rgb}{0.96,0.96,0.98}
\definecolor{clPurpleBg}{HTML}{E9F0F8} %
\definecolor{clPurple}{HTML}{C6CFDA}
\definecolor{clGreen}{RGB}{0,100,0}
\definecolor{clRed}{rgb}{0.8,0.25,0.33}
\newcommand{\cmark}{\textcolor{clGreen}{\checkmark}}
\newcommand{\xmark}{\textcolor{clRed}{\ding{55}}}
\newcommand{\grayrowcolor}{clGray}

\definecolor{ovRAG}{HTML}{3E405B}
\definecolor{ovRel}{HTML}{F9D487}
\definecolor{ovCo}{HTML}{E17A60}
\newcommand{\overviewbox}[3]{%
  \begin{tcolorbox}[width=#1\linewidth, colback=#2!12, colframe=#2!12, boxrule=0pt, arc=3pt,
    left=3pt, right=1pt, top=1.5pt, bottom=1.5pt, nobeforeafter, box align=top,
    fontupper=\scriptsize\raggedright]#3\end{tcolorbox}}
\newcommand{\ovmark}[1]{\makebox[1.1em][l]{#1}}

\definecolor{revision}{HTML}{53579c}

\newcommand\blfootnote[1]{%
  \begingroup
  \renewcommand\thefootnote{}\footnote{#1}%
  \addtocounter{footnote}{-1}%
  \endgroup
}

\newcommand{\aautoref}[1]{\hyperref[#1]{Appendix~\ref*{#1}}}

\newcommand{\eat}[1]{\ignorespaces}
\newcommand{\xxcomment}[4]{\textcolor{#1}{[$^{\textsc{#2}}_{\textsc{#3}}$ #4]}}

\newcommand{\citeit}[1]{\xxcomment{blue}{Y}{F}{citeit}}
\newcommand{\refit}[1]{\xxcomment{brown}{Y}{F}{refit}}

\definecolor{factbg}{rgb}{1.0, 0.93, 0.6}

\newcommand{\hlspan}[1]{{\setlength{\fboxsep}{0.5pt}\colorbox{factbg}{\strut #1}}}

\newcommand{\rellmlm}{\textsc{Rel-LMLM}\xspace}
\newcommand{\colmlm}{\textsc{Co-LMLM}\xspace}
\newcommand{\ours}{\colmlm}

\newcommand{\shortname}{\ours} %
\newcommand{\standard}{\textsc{Standard}}
\newcommand{\asker}{\textsc{LMLM-Asker}}

\newcommand{\lmlm}{\textsc{LMLM}}

\newcommand{\smollms}{\textsc{SmolLM2-135M}}
\newcommand{\smollmm}{\textsc{SmolLM2-360M}}

\newcommand{\hfsmollm}{\textsc{HF/SmolLM2}}
\newcommand{\hfsmollms}{\textsc{HF/SmolLM2-135M}}
\newcommand{\hfsmollmm}{\textsc{HF/SmolLM2-360M}}
\newcommand{\hfsmollml}{\textsc{HF/SmolLM2-1.7B}}

\newcommand{\finewebedu}{\textsc{FineWeb-Edu}}

\title{Co-LMLM: Continuous-Query Limited Memory Language Models}

\iclrfinalcopy
\author{%
  Yair Feldman\textsuperscript{*} \hspace{.3em} Linxi Zhao \hspace{.3em} Nathan Godey \hspace{.3em} Dongyoung Go \hspace{.3em} Yilun Hua\vspace{.2em}\\
  \textbf{Kilian Q. Weinberger \hspace{.3em} Jennifer J. Sun \hspace{.3em} Yoav Artzi}\vspace{.3em}\\
  Department of Computer Science\\
  Cornell University\vspace{.3em}\\
  \texttt{yairf@cs.cornell.edu}\\
  \texttt{\{lz586, ng554, dg793, yh2228\}@cornell.edu}\\
  \texttt{\{kilian, jennifer.sun, yoavartzi\}@cornell.edu}
}

\begin{document}

\maketitle
\lhead{Preprint. Under review.}   %
\ificlrfinal
\blfootnote{\textsuperscript{*}Individual author contributions are detailed in the \hyperlink{contributions}{acknowledgments}.}
\fi

\begin{abstract}

\input{sections/00-abstract}

\end{abstract}

\input{sections/01-introduction}

\input{sections/03-related_work}

\input{sections/04-method}

\input{sections/05-experiments}

\input{sections/06-discussion}

\ificlrfinal
\input{sections/20-ack}

\fi

\clearpage
\bibliographystyle{iclr2027_conference}
\bibliography{custom}

\appendix

\input{sections/30-appendix}

\end{document}

%% file: sections/00-abstract.tex
Externalizing knowledge in LLM pre-training is a promising avenue to achieve higher performance at smaller scales, control knowledge use, and overall increase model transparency. We propose continuous-query limited memory language models (\colmlm), an LLM that interleaves flexible vector retrieval queries with next-token predictions, and is pre-trained to copy knowledge returned from the KB, rather than memorize it. \colmlm is pre-trained with a scalable approach that jointly trains a knowledge-externalizing LLM, induces its knowledge base, and learns an expressive continuous retrieval mechanism. Across pre-training at multiple model scales, \colmlm outperforms prior knowledge-externalizing and vanilla LLMs in both perplexity and factual precision. At 360M scale, this includes lower perplexity than models pre-trained on 40$\times$ more data, and SimpleQA-verified performance  that is in line with gpt-4o-mini and higher than Claude Sonnet 4.5.

%% file: sections/01-introduction.tex
\vspace{-10pt}
\section{Introduction}\label{sec:intro}
\vspace{-5pt}

Recently, there has been increasing interest in large language models (LLMs) that are pre-trained to externalize~\citep{zhao2025pre} or separate knowledge~\citep{ghosal2025memorizationsinksisolatingmemorization,ghosal2026nativelyunlearnablelargelanguage,pouransari2026pretraininghierarchicalmemoriesseparating}.
Knowledge externalization departs dramatically from how knowledge is usually treated in LLM pre-training, where knowledge is completely indistinguishable from linguistic and reasoning skills, both in its acquisition and representation in parameter space. 
The potential of externalization is significant. LLMs that do not store knowledge can be much smaller, and train much faster, simply because less is asked from them. 
Externalization and separation open clean avenues for unlearning with little or no utility costs, and if the pathways for using knowledge remain legible, to monitoring, auditing, and overall increased transparency of model function. 
Critically, when done during pre-training, externalization does not rely on brittle post-training, and stops the integration of knowledge into the parameters at the source.

We propose continuous-query limited memory language models (\colmlm), an LLM that interleaves vector retrieval queries with next-token predictions, and is pre-trained to copy knowledge returned from the knowledge base (KB), rather than memorize it. 
The KB stores vector keys and string values, with each entry grounded to a source document. 
This allows for expressive retrieval pathways, where pre-training has the full flexibility of neural representation to shape retrieval representations, while retrieved values remain legible and interpretable. 
\colmlm is pre-trained via a scalable approach to jointly train a knowledge-externalizing LLM, induce its KB, and learn an expressive continuous retrieval mechanism. 
Training follows the common scalable next-token-prediction pre-training, with two key differences: pre-processing the data to simulate knowledge retrieval and adding a retrieval objective, both designed for pre-training scales.

We evaluate \ours{} against \rellmlm~\citep{zhao2025pre}, a prior LMLM-class method that relies on relational KB and retrieval, and standard baselines trained without knowledge externalization. 
Across model scales, \ours{} achieves lower perplexity  and substantially improves factual precision over both \rellmlm{} and standard models, while preserving downstream NLU performance. 
At the 360M scale, \ours{} achieves lower perplexities than off-the-shelf HF/\smollmm{}, which is pre-trained on 40$\times$ more data, and SimpleQA-verified~\citep{haas2026simpleqaverifiedreliablefactuality} scores that are in line with gpt-4o-mini and higher than Claude Sonnet 4.5.
At the same time, \ours{} retains the controllability benefits of the \lmlm{} class: its external memory remains editable, human readable, auditable, and supports direct unlearning through KB operations. 

\ificlrfinal
Code, data, and models are available at: \url{https://lil-lab.github.io/co-lmlm-web}.
\else
Code, data, and models will be available upon publication.
\fi

\begin{figure}[t]
\centering
\includegraphics[width=\linewidth]{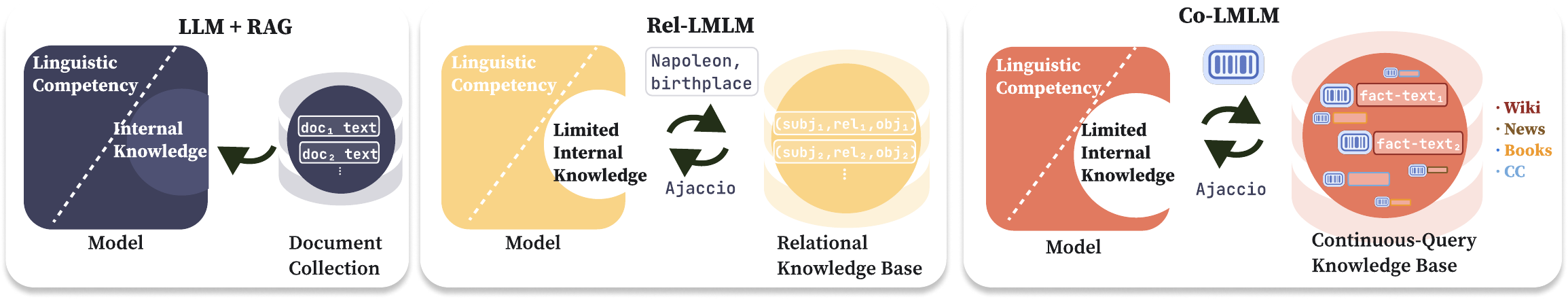}\\[1pt]
\hspace*{0.0036\linewidth}%
\overviewbox{0.2572}{ovRAG}{\ovmark{\xmark}Facts memorized in weights\\\ovmark{\cmark}Retrieves from any text\\\ovmark{\xmark}Separate retriever model}%
\hspace{0.0072\linewidth}%
\overviewbox{0.3357}{ovRel}{\ovmark{\cmark}Facts in an editable, readable KB\\\ovmark{\xmark}Relational facts, Wikipedia only\\\ovmark{\xmark}Multi-token decoded queries}%
\hspace{0.0108\linewidth}%
\overviewbox{0.3817}{ovCo}{\ovmark{\cmark}Facts in an editable, readable KB\\\ovmark{\cmark}Free-form facts from any text\\\ovmark{\cmark}Single-step continuous queries}%
\caption{\textbf{Knowledge separation across three regimes.} A standard LLM with RAG retrieves over external documents but keeps factual knowledge in its parameters (\emph{left}); \rellmlm{} externalizes facts to a relational KB queried with an explicit decoded query (\emph{middle}); \ours{} externalizes facts to an unstructured index, retrieved directly from the model's hidden state (\emph{right}). \emph{Bottom:} some of the advantages and limitations of each regime.}
\vspace{-1em}
\label{fig:overview}
\end{figure}

%% file: sections/03-related_work.tex
\section{Related Work}\label{sec:related}

\paragraph{Parametric Knowledge and its Limitations}
Pre-trained language models store factual knowledge implicitly in their parameters and recall it at generation time~\citep{petroni2019languagemodelsknowledgebases, roberts2020much}. 
This form of storage is inherently limited: factual recall scales with model size and exposure frequency, making long-tail knowledge difficult to capture~\citep{kandpal2023largelanguagemodelsstruggle, mallen-etal-2023-trust}. 
Moreover, factual associations are entangled with linguistic representations, making individual facts difficult to attribute, edit, or remove, and contributing to hallucinations and stale knowledge~\citep{elhage2022toy, dai2022knowledge, huang2024advancinglargelanguagemodel}. 
They also make \textit{knowledge editing} and \textit{unlearning} challenging: removing or modifying specific facts typically requires additional training or specialized objectives, often trading off with general capabilities~\citep{maini2024tofutaskfictitiousunlearning}. 
MemSinks isolates memorization-specific parameters, but still keeps knowledge within the model~\citep{ghosal2025memorizationsinksisolatingmemorization}.
These challenges motivate approaches that move factual knowledge into separate, editable stores.

\paragraph{Externalizing Knowledge from Model Weights}
A growing line of work augments language models with non-parametric or semi-parametric memory, or aims to move factual knowledge more explicitly \emph{out of model parameters}.
kNN-LM retrieves nearest-neighbor examples from an external datastore and interpolates their probabilities with the model distribution~\citep{khandelwal2020generalizationmemorizationnearestneighbor}, while RETRO retrieves neighboring chunks and conditions on them through cross-attention~\citep{borgeaud2022improvinglanguagemodelsretrieving}. 
Other approaches introduce sparse memory layers or learned key-value memories to increase model capacity and factuality~\citep{lample2019large, berges2024memorylayers, wu2022memorizingtransformers, peng2023semiparametriclanguagemodelsscalable, yang2024text, cheng2026conditionalmemoryscalablelookup,tseng2026l3largelookuplayers}. 
These methods can improve perplexity or factuality, but their memory is not directly inspectable or editable. 

\paragraph{Limited Memory Language Models}
\citet{zhao2025pre} is the most related work to ours. 
\autoref{fig:overview} illustrates some of the similarities and differences. 
They introduced Limited Memory Language Models (LMLMs), which externalize factual knowledge into a textual relational KB during pre-training rather than encoding it in model parameters. We refer to this approach as \rellmlm to distinguish it from our continuous-query formulation. \rellmlm enables faster pre-training, interpretable retrieval, and direct editing, but its relational design limits the external memory coverage: facts must be expressible as Wikipedia-style subject-relation-object tuples, and retrieval depends on brittle textual queries that introduce additional generation token overhead. 
\rellmlm training relies on data pre-processing to simulate KB queries, and otherwise follows conventional next-token-prediction pre-training. 
Although scaleable, it limits the expressivity of retrieval pathways, as the simulated relational queries dictate the exact form of querying.
\colmlm preserves the benefits of an externalized, editable KB while relaxing these constraints. It replaces relational tuples with free-form textual knowledge spans, and decoded relational queries with continuous queries emitted from the model's hidden states. These queries can carry richer contextual information for retrieval, allowing \colmlm to scale beyond Wikipedia-style relational facts to general corpora (e.g., \finewebedu{}), while reducing the token overhead of query generation.
The \colmlm pre-training process is rooted at the pre-processing approach of \rellmlm, but avoids from dictating retrieval representations, instead leaving these for learning. 

\paragraph{Retrieval-Augmented Language Models}
Retrieval-Augmented Generation (RAG), also illustrated in \autoref{fig:overview}, conditions generation on retrieved passages during inference~\citep{lewis2021retrievalaugmentedgenerationknowledgeintensivenlp, izacard2021leveragingpassageretrievalgenerative}, while retrieval-aware pre-training methods incorporate retrieval into model training~\citep{guu2020realmretrievalaugmentedlanguagemodel, borgeaud2022improvinglanguagemodelsretrieving, izacard2022atlasfewshotlearningretrieval}. 
Other methods let models decide when to retrieve and issue textual queries, as in Self-RAG, DRAGIN, and LMLM~\citep{asai2023selfraglearningretrievegenerate, su2024dragin, zhao2025pre}.
Our technique is related to methods that generate retrieval signals directly from internal representations rather than decoded queries~\citep{muennighoff2024generative, zhang2024onegen, zhang2025imprag}. 
Unlike most retrieval-augmented methods, which add external context while leaving parametric knowledge largely intact, \colmlm uses retrieval to define the parametric/non-parametric boundary during pretraining: factual spans are externalized into a schema-free index, and the model learns when and how to retrieve them through a contrastive objective trained jointly with next-token prediction.

%% file: sections/04-method.tex
\section{Method}\label{sec:method}

\begin{figure}[t]
\centering
\includegraphics[width=\linewidth]{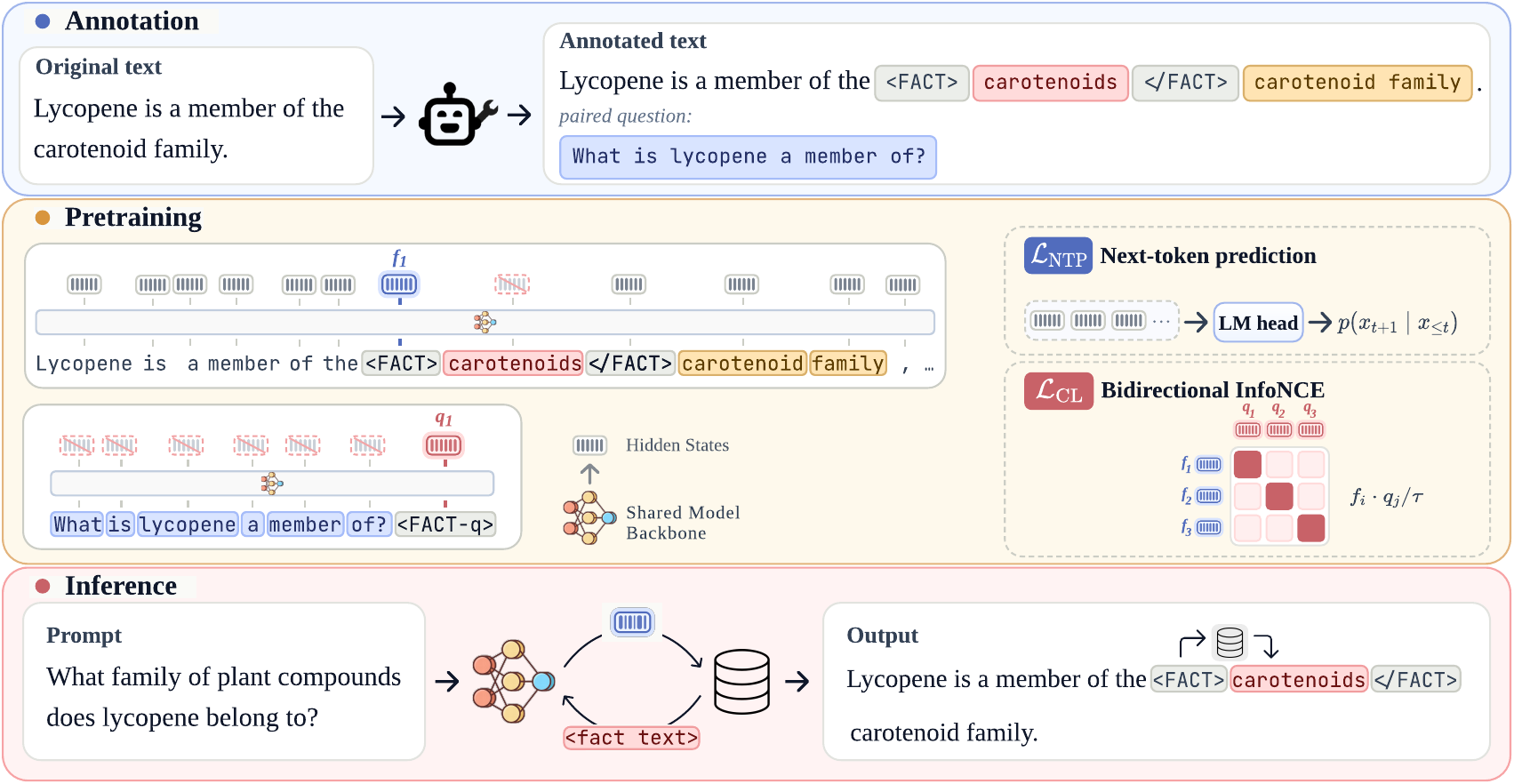}
\caption{\textbf{Overview of \ours{}.}
\emph{Top:} We annotate the pre-training corpus with fact spans and paired questions, while leaving the document text unchanged.
\emph{Middle:} During training, a single decoder-only Transformer processes both the annotated document and its paired questions. We optimize next-token prediction over non-fact tokens and a bidirectional InfoNCE loss that aligns the query representation at each \texttt{<FACT>} token with the question representation at the corresponding \texttt{<FACT-q>} token.
\emph{Bottom:} At inference, emitting \texttt{<FACT>} triggers dense retrieval using the hidden state at the emitted \texttt{<FACT>} position as the query. The retrieved fact span is inserted into the context, followed by \texttt{</FACT>}, and decoding continues. No textual query is decoded.}
\label{fig:method-overview}
\vspace{-10pt}
\end{figure}

The \ours{} model is a decoder-only Transformer, which functions as both a language model and a dense retriever. 
\autoref{fig:method-overview} provides a high-level overview of the method.
The model generates query vectors similar to tool calling, and the returned values are encoded by the model, before decoding resumes (\autoref{sec:method:modeling}). 
We jointly train the language modeling and retriever functions (\autoref{sec:method:learning}), using pre-training data that is pre-processed to simulate retrieval, paired with synthetic questions that provide supervision for the contrastive retrieval objective (\autoref{sec:method:annotation}).

\subsection{Modeling and Inference}\label{sec:method:modeling}

\ours{} performs retrieval through continuous tool calls.
The external memory is a key-value store: each key is a dense vector produced by the model, and each value is a factual text span.
Unlike standard tool-calling methods~\citep{schick2023toolformer}, the retrieval query is not represented as text.
Instead, the query is the model's hidden state at a special token.

At inference time, the model decodes tokens autoregressively. When it emits the special \texttt{<FACT>} token, we read the last-layer hidden state at that position as a retrieval query, query the index, and retrieve the top-1 fact text snippet. We splice the retrieved snippet into the context immediately after \texttt{<FACT>}, append a closing \texttt{</FACT>} special token, and resume autoregressive decoding from \texttt{</FACT>} onward. Each retrieval therefore costs one extra autoregressive step (the opening \texttt{<FACT>} token) and grows the context by the length of the retrieved span.

\newcommand{\featfunc}{\phi}
\newcommand{\question}{Q}

\subsection{Training}\label{sec:method:learning}

Our training data consists of raw text in which factual spans are enclosed by \texttt{<FACT>} and \texttt{</FACT>}.
Each marked span is paired with a natural-language question whose answer is the span.
\autoref{fig:method-overview} (top) illustrates a single example.
\autoref{sec:method:annotation} describes how we pre-process raw pre-training data to create this data. 
Our objectives are to train the model to both predict the next token and externalize the information in the delimited spans, construct a KB from all such spans, and to emit retrieval queries to fetch spans from the KB as needed. 
The key to training the retrieval function is contrastive query pairs between documents and questions, allowing for both positive and negative pairs.

Let $x$ be an input document with factual spans bracketed by \texttt{<FACT>} and \texttt{</FACT>}. For each factual span, we have a question $\question$. The question is synthesized so its answer is the corresponding bracketed span in $x$, and it is appended with a query marker token \texttt{<FACT-q>}. 
The retrieval vectors at the query markers appended to the questions are used as positive examples for the contrastive retrieval loss. 
The pre-training panel of \autoref{fig:method-overview} (middle) visualizes how $x$ and $\question$ are used.

Let $p(x_t \mid x_{<t}; \theta)$ be the probability of the token $x_t$ in the sequence document $x$, parameterized by $\theta$. 
In our model, $p_\theta$ is computed by a Transformer. 
We denote $\featfunc_t(x_{<t}; \theta)$ as the features computed by the Transformer for index $t$ in sequence $x$.\footnote{We L2-normalize the feature representations as they are used as retrieval queries~\citep{simclr2020}.} This is the same Transformer used for $p_\theta$. 

We jointly minimize next-token prediction (NTP) and bidirectional contrastive losses. 
The NTP loss is:
$\mathcal{L}_{\mathrm{NTP}}(\theta)=-\sum_{t\notin M}\log p(x_t\mid x_{<t};\theta)$,
where $M$ is the set of positions in $x$ that follow an opening \texttt{<FACT>}, up to and including its closing \texttt{</FACT>}. 
This NTP formulation means that we do not optimize our model to memorize the bracketed fact tokens, which represent the knowledge we aim to externalize. This is a critical design choice to externalize knowledge. 
We do optimize the loss of the opening \texttt{<FACT>} token, because the model must know to generate it in order to trigger retrieval. The closing \texttt{</FACT>} is appended mechanically, so it does not require optimization.

The contrastive loss treats each fact query from the document $x$ and its corresponding question $\question$ as a positive pair, and all other questions as negatives. 
For a fact that starts at index $t$ in a document $x$, let $f = \featfunc_t(x_{<t}; \theta)$ be the query representation for retrieving this fact, i.e., the feature representation for the \texttt{<FACT>} token opening this fact span. 
Correspondingly, for a question $\question$ where the query marker \texttt{<FACT-q>} is at index $t$, let $q = \featfunc_t(\question_{<t}; \theta)$ be the question representation, i.e., the feature representation computed for \texttt{<FACT-q>}.
We use the InfoNCE loss~\citep{oord2019representationlearningcontrastivepredictive}, computed on the current batch.
Let $\mathcal{B} = \{(f^{(i)}, q^{(i)})\}_{i=1}^{B}$  be the set of all paired fact queries $f^{(i)}$ and question queries $q^{(i)}$ in the current batch. 
For each $(f^{(i)}, q^{(i)}) \in \mathcal{B}$, the loss terms are:
\begin{small}
\begin{equation}
\ell_{f \to q}^{(i)} \;=\; \log \frac{\exp(f^{(i)} \cdot q^{(i)} / \tau)}{\sum_{k=1}^{B} \exp(f^{(i)} \cdot q^{(k)} / \tau)}
\qquad
\ell_{q \to f}^{(i)} \;=\; \log \frac{\exp(q^{(i)} \cdot f^{(i)} / \tau)}{\sum_{k=1}^{B} \exp(q^{(i)} \cdot f^{(k)} / \tau)},
\end{equation}
\end{small}%
where $\tau$ is a temperature hyperparameter. 
We average the two directions:
\begin{small}
\begin{equation}
\mathcal{L}_{\mathrm{CL}} \;=\; -\frac{1}{2B} \sum_{i=1}^{B} \big( \ell_{f \to q}^{(i)} \,+\, \ell_{q \to f}^{(i)} \big).
\end{equation}
\end{small}
We optimize the joint loss $\mathcal{L} = \mathcal{L}_{\mathrm{NTP}} + \lambda\mathcal{L}_{\mathrm{CL}}$, with $\lambda$ a hyperparameter set to $0.25$ by default.

\paragraph{Building the Index}

We build a dense retrieval index from our pre-training data. The index pairs vector keys with fact spans. 
We run the pre-trained language model over the pre-training corpus, extract the L2-normalized hidden state $\featfunc(\cdot ; \theta)$ at every \texttt{<FACT>} position, and store it in a dense vector index as a key whose value is the verbatim in-text span at that position. Indexing a new corpus therefore requires only annotating its fact spans (\autoref{sec:method:annotation}) and a forward pass of \ours{}.

\begin{figure}[t]
\centering
\includegraphics[width=\linewidth]{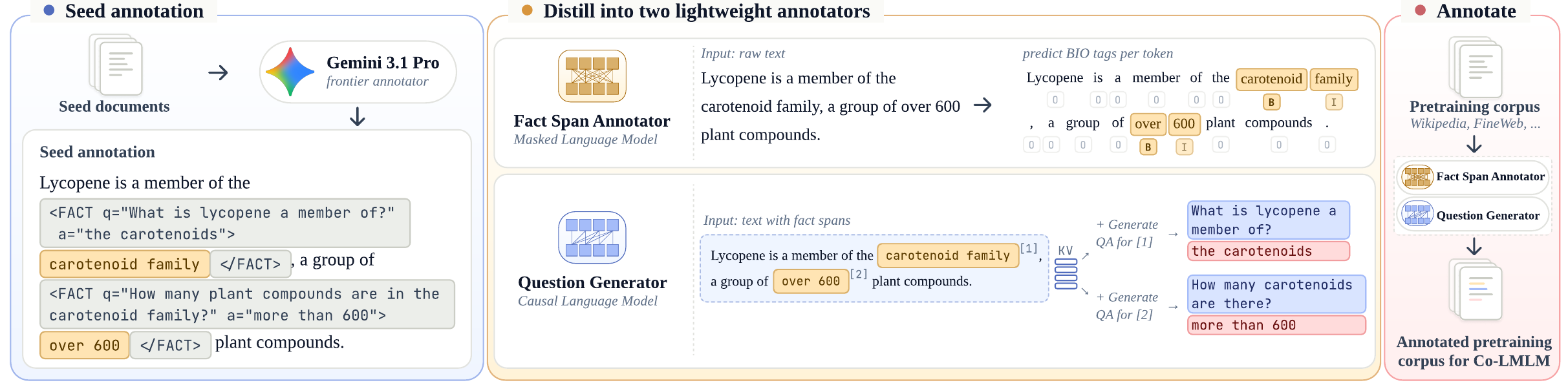}
\caption{\textbf{Annotation pipeline.}
\emph{Left:} A frontier model annotates a small seed set of documents in the \texttt{<FACT q=...\ a=...>span</FACT>} format.
\emph{Middle:} We use the seed set to train two lightweight annotators: a fact span annotator that BIO-tags fact spans in raw text, and a question generator that produces a question-answer pair for each span; the document prefix is KV-cached and reused across all spans.
\emph{Right:} We run the two annotator over the full pre-training corpus, converting raw documents into the annotated pre-training corpus for \ours{}.}\label{fig:annotation-pipeline}
\vspace{-10pt}
\end{figure}

\subsection{Training Data Pre-processing}\label{sec:method:annotation}

Training requires documents annotated with fact spans and paired questions (\autoref{sec:method:learning}). 
We design an automated, scalable process to annotate raw pre-training data. 
The pre-processing pipeline has three stages: annotate a seed set by prompting a state-of-the-art frontier LLM; fine-tune lightweight annotators on the seed data; and apply them to the full pre-training corpus. 
Critically, we fine-tune smaller models because of the cost of annotating at scale with a large frontier model.

\paragraph{Annotation Specification}

We annotate each training document with a set of \emph{fact spans}. A fact span is a substring of a pre-training document $x$ that encodes world knowledge we want our model to externalize --- to retrieve rather than memorize. Each fact span is associated with two additional pieces of information: a context-dependent question that captures the retrieval intent at this position without itself carrying any factual content, and a snippet-form answer that answers the question in a form that need not match the surface form of the span. The question is used as the question $\question$ during training. 
An annotated fact takes the form \texttt{<FACT q=\{question\} a=\{answer\}>\{span\}</FACT>}. 

The annotation reflects three guidelines: 
(a) spans are the narrowest self-contained substrings that fully answer the question; 
(b) questions are written as standalone search queries that name their subject explicitly, and they avoid mentioning information that appears later in the document so that no future fact leaks through the question;
and (c) answers vary the surface form of the span, so the model is exposed at training time to retrieved values that differ lexically from the surrounding context. We provide more details about the prompt used for the seed annotation in \aautoref{app:gemini-prompt}.

\paragraph{Annotation Pipeline}

\autoref{fig:annotation-pipeline} visualizes the annotation pipeline. 
We first use Gemini-3.1-Pro to annotate a small seed set of documents using the \texttt{q="..."}/\texttt{a="..."} format defined above. The seed set is then used to distill two efficient annotators for the entire pre-training corpus. The \emph{fact span annotator} is a masked language model that detects fact spans on the original text via BIO tagging,\footnote{BIO stands for begin-inside-outside~\citep{ramshaw-marcus-1995-text}.} built on a ModernBERT encoder~\citep{warner-etal-2025-smarter}. The \emph{question generator} is a decoder-only generator that, conditioned on the document with span markers inserted, emits a question-answer pair for each tagged span; the document context is encoded once and reused across all facts. We assess the quality of the annotation pipeline in \aautoref{app:annotation_quality}.

\paragraph{Why Two Annotators?}
A single end-to-end generator that produces the full annotated text in one shot is the obvious alternative. We chose the split design for three reasons. First, for efficiency reasons: the question generator only emits questions and answers, not the document text, which lets us prefill the document context once and reuse it across all of its facts. Second, predicting spans as offsets into the original text makes the annotations \emph{faithful by construction}: the in-text span is always a verbatim substring of the source document, while a generative annotator can drift away from the source (i.e., hallucinate), especially on long documents.
Finally, indexing new data is highly efficient as it only involves two forward passes (one for the span annotator and one for \ours{}).

%% file: sections/05-experiments.tex
\section{Experiments}\label{sec:experiments}

\subsection{Experimental Setup}

We train \ours{} using the \smollms{} and \smollmm{} architectures~\citep{allal2025smollm2smolgoesbig}, with a standard language model trained at each size as the \standard{} baseline. To facilitate a fair and controlled comparison with \rellmlm{}, we train two \rellmlm{} models based on the same architectures and use a high-quality Wikipedia corpus (${\sim}3\text{B}$ tokens)\footnote{\url{https://huggingface.co/datasets/allenai/dolmino-mix-1124}}, following \citet{zhao2025pre}. We train all Wikipedia models for 100K steps with a context length of 4{,}096 tokens.  
We show the generalization of our process to general web text by training \ours{} and a corresponding \standard{} baseline on 90B tokens sampled from \finewebedu{}~\citep{penedo2024finewebdatasetsdecantingweb}.
We experiment with two KBs.
We create a KB from Wikipedia only for a direct comparison with the \rellmlm{} KB that is constructed from this same corpus. It also guarantees that the KB does not cover information  unavailable to the baselines. This the default KB, unless another is specified. 
The \colmlm Wikipedia KB includes ${\sim}240\text{M}$ items, compared to \rellmlm's ${\sim}145\text{M}$, demonstrating the less restricted, and more aggressive externalization of \colmlm. 
In our scaling-up experiments, we construct the KB from our \finewebedu{} corpus as well (${\sim}2.2\text{B}$ items). 
\aautoref{sup:exp_setting} provides additional implementation details.

\subsection{Language Modeling Perplexity}
\label{sec:ppl}

\paragraph{Evaluation Setup}

We evaluate using perplexity on 1{,}000 randomly sampled Simple English Wikipedia (Simple-Wiki) articles.\footnote{\url{https://simple.wikipedia.org/}}
These articles are not included in our pre-training corpus, but their underlying facts are likely covered by our index in different surface forms, because the information in Simple-Wiki is largely a subset of Wikipedia.
This setup avoids coverage issues and helps distinguish retrieval or annotation errors in \ours{} from imperfect parametric memorization in \standard{}. 
The key comparison is between the models we train, controlling for the training data. We also include off-the-shelf models, with the caveat of a potential Wikipedia domain shift.

\newcommand{\pplfigscale}{0.54}

\begin{figure*}[t]
\centering

\begin{subfigure}[t]{0.46\textwidth}
  \centering
  \includegraphics[scale=\pplfigscale]{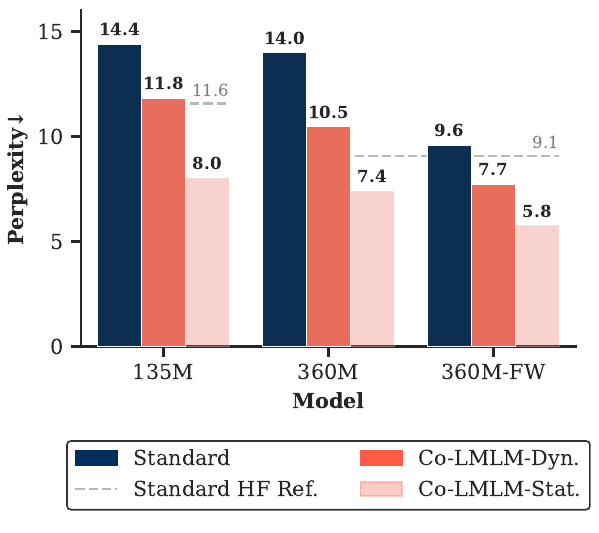}
  \caption{Controlled Wikipedia perplexities}\label{fig:eval_ppl_sizes}
\end{subfigure}
\hfill
\begin{subfigure}[t]{0.52\textwidth}
  \centering
  \includegraphics[scale=\pplfigscale]{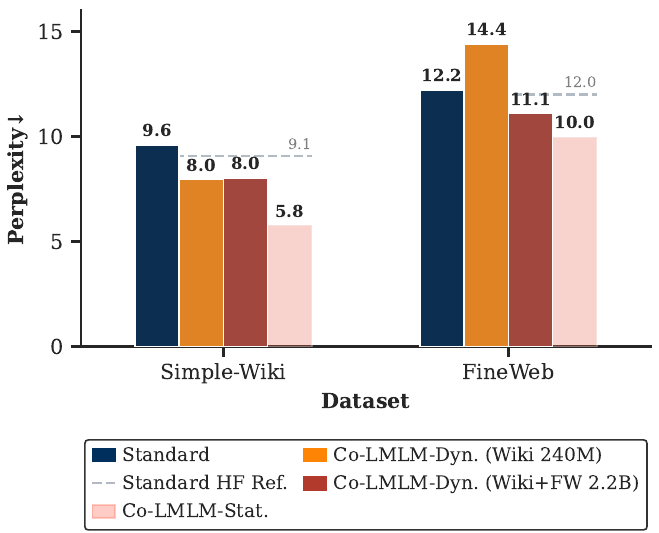}
  \caption{Data scaling-up perplexities}\label{fig:eval_ppl_index}
\end{subfigure}

\caption{\textbf{Perplexity evaluation.}
(\subref{fig:eval_ppl_sizes}) Simple-Wiki test perplexity comparison between \standard{} and \ours{} across two model sizes (135M and 360M) and two training corpora (Wikipedia only vs.\ Wikipedia and \finewebedu{}). 
(\subref{fig:eval_ppl_index}) Perplexities of \standard{}-360M-FW and \ours{}-360M-FW on Simple-Wiki and FineWeb test splits under different KB index scales --- 240M items from Wikipedia vs.\ 2.2B from Wikipedia and \finewebedu{}. We show two perplexity variants for \ours{} (dynamic and static). The dashed gray line reports the perplexity of the same-size \hfsmollm{} model for context. The HF models are trained on much more data. The left figure uses PQ240 quantization, while the right uses PQ96 for both indices.}\label{fig:ppl_evals}
\end{figure*}

\input{tables/table_factuality}

\newcommand{\evalfigscale}{0.6}

\begin{figure*}[t]
\centering

\begin{subfigure}[t]{0.34\textwidth}
  \centering
  \includegraphics[scale=\evalfigscale]{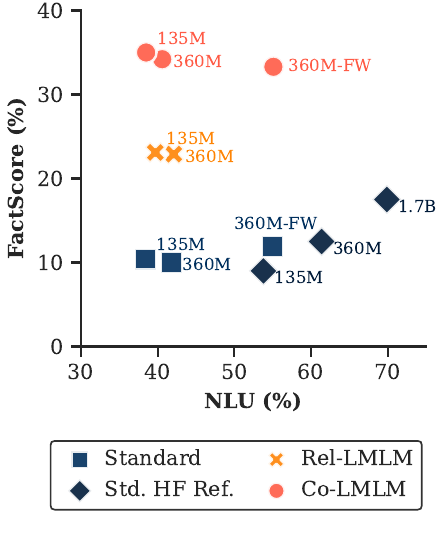}
  \caption{Factuality vs.\ NLU}\label{fig:eval_nlu_fact}
\end{subfigure}
\hfill
\begin{subfigure}[t]{0.65\textwidth}
  \centering
  \includegraphics[scale=\evalfigscale]{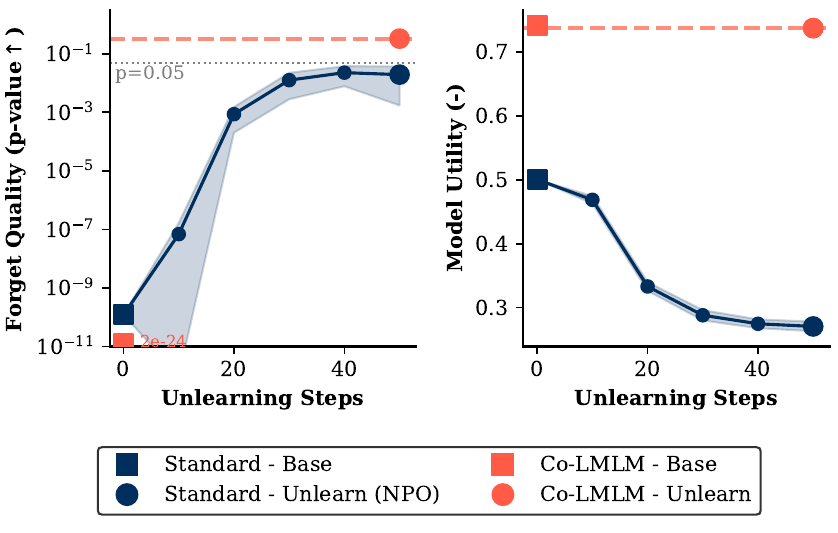}
  \caption{Unlearning on TOFU}\label{fig:eval_tofu}
\end{subfigure}

\caption{\textbf{Factuality, NLU, and unlearning evaluation.}
(\subref{fig:eval_nlu_fact}) FactScore and NLU tradeoff for various models --- \ours{} substantially improves factual precision while preserving NLU performance comparable to \standard{} models.
(\subref{fig:eval_tofu}) Machine unlearning on TOFU ---      \ours{} forgets through direct KB operations, without additional training or utility degradation.}
\label{fig:general_eval_summary}
\vspace{10pt}
\end{figure*}

We report two perplexity variants, following~\citet{zhao2025pre}. 
\emph{Static (Oracle)} assumes perfect lookup behavior by supplying the correct retrieved spans as context, giving an optimistic estimate of language modeling quality.
\emph{Dynamic} performs actual lookups during decoding, so retrieval performance is reflected in the score.
While dynamic follows the standard perplexity calculation, it introduces some complexities in our case. We discuss this in depth in \aautoref{app:eval_benchmark} and provide a worst-case analysis. Importantly, this worst-case analysis does not change our conclusions.

\paragraph{Results}

\looseness=-1 
\ours{} consistently achieves lower perplexity than \standard{} across model sizes and perplexity variants (\autoref{fig:ppl_evals}).
\autoref{fig:eval_ppl_sizes} shows perplexities for Simple-Wiki, where we are fairly certain to get full coverage of factual data. 
Under the more realistic \emph{Dynamic} setting, \ours{} reduces perplexity from 14.4 to 11.8 for 135M and from 14.0 to 10.5 for 360M.
These gains suggest that our annotated pretraining data and training objective successfully improve language modeling over factual text.
The FineWeb (FW) results indicate the benefit of data scale, charting initial scaling trends. 
Off-the-shelf models (HF/\smollms{} and HF/\smollmm{}, dashed line) provide context: the relative gap \colmlm-360M-FW shows is particularly promising given the orders of magnitude more training data of the HF models, although a likely domain confounder qualifies this result. 

\autoref{fig:eval_ppl_index} provides our main perplexity result, scaling up the index to include the entire FW and Wikipedia corpora used for training our models. This reflects a realistic KB construction, both in document diversity and size. 
This results in a much larger index of 2.2B entries, compared to the 240M entries in the Wiki-only index. 
We report results on both the Simple-Wiki dataset and an 895-document held-out test set from the FW dataset. 
We compute these results with higher quantization for an additional 2.5$\times$ memory reduction per vector, so all indices fit in memory. 

We do not expect benefits from scaling up the index to include FW data for Simple-Wiki, because the corresponding knowledge is expected to be contained in the English Wikipedia dataset. Indeed, under the same quantization level, we only see a minor increase in perplexity even though the index is one order of magnitude larger, which speaks to the robustness of our learned retriever. 
For the FW test set, on the other hand, we observe a substantial improvement over the Wiki-only index, producing a significantly lower perplexity than that of the HF/\smollmm{} model, which is pre-trained on those documents and overall much more data (4T tokens). The poor performance of the Wiki-only index on the FW test set demonstrates Wikipedia's limited coverage of knowledge and the need to scale up the factual KB. This result demonstrates effective scaling beyond Wikipedia data.

\subsection{Factuality and General Benchmark Performance}
\label{sec:factuality}

\paragraph{Evaluation Setup} 
We evaluate factuality across five complementary settings that probe different forms of factual knowledge in pre-trained models.
TriviaQA~\citep{joshi2017triviaqalargescaledistantly} tests broad open-domain QA, while PopQA~\citep{mallen-etal-2023-trust} focuses on long-tail entities. 
SimpleQA Verified~\citep[hereafter SimpleQA;][]{wei2024measuringshortformfactualitylarge, haas2026simpleqaverifiedreliablefactuality} targets short-form, fact-seeking questions that are deliberately difficult and have a single, indisputable answer.
T-REx~\citep{elsahar-etal-2018-rex} evaluates controlled factual completion, where the model completes a factual statement with the correct object entity. FactScore~\citep{min2023factscorefinegrainedatomicevaluation} evaluates long-form factual precision in biography generation by decomposing generations into atomic facts and verifying them against external evidence.
We also evaluate \ours{} on standard NLU benchmarks to verify that factual offloading does not compromise language understanding.

\paragraph{Results}
\autoref{tab:eval_factual_precision} shows that \ours{} consistently improves over the corresponding \standard{} baselines, with especially large gains on short-form QA and FactScore, exceeding $18$ points in each case. 
As with perplexity, we include off-the-shelf models for contextualization, with similar caveats. 
\ours{} also outperforms \rellmlm{}, demonstrating that continuous queries are more robust than textual entity-relation queries.
We attribute these gains to two factors: free-form factual spans provide richer external knowledge than relational triplets, and hidden-state queries can use the full generation context rather than a short and restricted decoded query.
Training on FineWeb-Edu further improves short-form QA and knowledge completion, showing that our design remains effective beyond Wikipedia-style pre-training data. 
\ours-360M-FW achieves 21.7 on SimpleQA, in line with gpt-4o-mini and above Claude Sonnet 4.5 and Grok 2 according to the public leaderboard.\footnote{\url{https://www.kaggle.com/benchmarks/deepmind/simpleqa-verified}}

\ours{} achieves a better factuality--NLU trade-off than standard models, with much higher factuality and similar NLU capabilities (\autoref{fig:eval_nlu_fact}). \aautoref{app:nlu_details} details NLU results across tasks.
We also compare \ours{} to controlled RAG baselines in \aautoref{app:rag_comparison} (\autoref{tab:comparison_lmlm_rag_smollmm}), finding that RAG and \shortname{} are complementary and that \ours{}+RAG is a viable path forward.

\subsection{Unlearning}
\label{sec:unlearning}

Externalizing knowledge simplifies unlearning: instead of retraining the model, we just remove the relevant entries from memory. 
We evaluate this property with TOFU~\citep{maini2024tofutaskfictitiousunlearning}, following the same setup as \rellmlm{} (\aautoref{sup:tofu-details}). TOFU tests whether a model can forget a designated \emph{forget set} while preserving general utility. Its main metric, \emph{forget quality}, tests whether the unlearned model is statistically indistinguishable from a retain-only model on the forgotten examples.
\autoref{fig:eval_tofu} shows how simple memory operations achieve effective forgetting ($p$-value $> 0.05$) while preserving model utility. In contrast, training-based unlearning methods such as NPO~\citep{zhang2024negativepreferenceoptimizationcatastrophic} update model parameters and sacrifice utility due to entangled parametric knowledge. 
\ours{} retains the controllability benefit of \rellmlm{} while being more expressive and efficient.

\subsection{Knowledge Memorization}

\input{tables/table_factuality_wo_db}

To measure how much factual performance depends on access to the learned external memory, we evaluate a no-retrieval variant of our model. In this setting, we disable the model's ability to trigger KB retrieval during generation. Concretely, for continuous-query models, we apply a logit bias of $-100$ to all factual special tokens that initiate or delimit retrieval spans. This prevents the model from emitting the markup required to invoke the retrieval loop, forcing it to generate plain text from its parameters alone. This ablation isolates the contribution of external memory access from the model's parametric knowledge.
\autoref{tab:ablation_disable_database} shows that factual precision drops substantially. This shows that \ours{} genuinely shifts factual knowledge from model weights to the external memory.

\subsection{Additional Experiments to Analyze the Query Mechanism}\label{subsec:lmlm-asker}

We isolate the impact of using continuous queries by comparing \colmlm to \textsc{LMLM-Asker}, a variant that generates free-form natural-language questions as queries.
\autoref{tab:lm_asker_comparison} in the appendix shows that \ours{} outperforms \textsc{LMLM-Asker}, demonstrating  the advantage of continuous queries. 
Text queries also add generation cost: \textsc{LMLM-Asker} and \rellmlm{} decode a question or a relational query, respectively, before every lookup, while \ours{} reads the query from its hidden state, saving a cost linear in the length of the question (\aautoref{app: latency}).
We observe that deciding \emph{when} to retrieve limits \ours{} performance: forcing a lookup before answering, while keeping the model's own query, improves scores by up to 9.3 points (\aautoref{sec:eval:query_timing}). An error analysis of \ours{}-360M-FW (\aautoref{app:error_analysis}) finds that most of its PopQA errors stem from incorrect retrievals, fewer from missing ones, and only rarely from misusing a retrieved fact.

%% file: tables/table_factuality.tex
\begin{table}[t]
\centering
\caption{\textbf{Factual precision evaluation.} Green subscripts show the absolute gain over the respective \standard{} baseline. We include * models for context, although they are trained on much larger corpora (2T/4T/11T tokens for 135M/360M/1.7B models).
Our rebuilt \rellmlm{} baseline follows the original setup with engineering improvements for a stronger, fairer comparison (\aautoref{app:rellmlm_reimplementation}).}
\label{tab:eval_factual_precision}
\scriptsize
{\setlength{\tabcolsep}{10pt}
\renewcommand{\arraystretch}{1.08}
\resizebox{1.0\textwidth}{!}{%
\begin{tabular}{lccccc}
\toprule
\multirow{2}{*}{Model}
& \multicolumn{3}{c}{Short-form QA}
& \multicolumn{1}{c}{Knowledge Completion}
& \multicolumn{1}{c}{Long-form Generation} \\
\cmidrule(lr){2-4}\cmidrule(lr){5-5}\cmidrule(lr){6-6}
& TriviaQA ↑ & PopQA ↑ & SimpleQA ↑ & T-REx EM ↑ & FactScore ↑ \\
\midrule

$\hfsmollms$$^{*}$
& 14.7 & 18.9 & 1.3 & 36.1 & 9.0 \\

\textsc{\standard{}-135M}
& 6.4 & 15.2 & 0.1 & 38.2 & 10.4 \\

\textsc{Rel-LMLM-135M}
& 13.4 & 27.2 & 6.2 & 42.9 & 23.1 \\

\cellcolor{clPurpleBg}\textbf{\textsc{Co-LMLM-135M}}
& \cellcolor{clPurpleBg}28.4{\rlap{\smash{\textcolor{clGreen}{$_{+22.0}$}}}}
& \cellcolor{clPurpleBg}47.3{\rlap{\smash{\textcolor{clGreen}{$_{+32.1}$}}}}
& \cellcolor{clPurpleBg}18.2{\rlap{\smash{\textcolor{clGreen}{$_{+18.1}$}}}}
& \cellcolor{clPurpleBg}40.5{\rlap{\smash{\textcolor{clGreen}{$_{+2.3}$}}}}
& \cellcolor{clPurpleBg}35.0{\rlap{\smash{\textcolor{clGreen}{$_{+24.6}$}}}} \\

\midrule

$\hfsmollmm$$^{*}$
& 25.4 & 22.7 & 1.4 & 45.8 & 12.5 \\

\textsc{\standard{}-360M}
& 7.5 & 16.2 & 0.3 & 46.3 & 10.0 \\

\textsc{\standard{}-360M-FW}
& 18.6 & 18.9 & 1.3 & 50.5 & 11.9 \\

\textsc{Rel-LMLM-360M}
& 15.6 & 26.6 & 5.6 & 47.4 & 22.9 \\

\cellcolor{clPurpleBg}\textbf{\textsc{Co-LMLM-360M}}
& \cellcolor{clPurpleBg}31.4{\rlap{\smash{\textcolor{clGreen}{$_{+23.9}$}}}}
& \cellcolor{clPurpleBg}46.5{\rlap{\smash{\textcolor{clGreen}{$_{+30.3}$}}}}
& \cellcolor{clPurpleBg}21.2{\rlap{\smash{\textcolor{clGreen}{$_{+20.9}$}}}}
& \cellcolor{clPurpleBg}47.4{\rlap{\smash{\textcolor{clGreen}{$_{+1.1}$}}}}
& \cellcolor{clPurpleBg}34.2{\rlap{\smash{\textcolor{clGreen}{$_{+24.2}$}}}} \\

\addlinespace[1 pt]
\cellcolor{clPurpleBg}\textbf{\textsc{Co-LMLM-360M-FW}}
& \cellcolor{clPurpleBg}36.9{\rlap{\smash{\textcolor{clGreen}{$_{+18.3}$}}}}
& \cellcolor{clPurpleBg}50.6{\rlap{\smash{\textcolor{clGreen}{$_{+31.7}$}}}}
& \cellcolor{clPurpleBg}21.7{\rlap{\smash{\textcolor{clGreen}{$_{+20.4}$}}}}
& \cellcolor{clPurpleBg}54.5{\rlap{\smash{\textcolor{clGreen}{$_{+4.0}$}}}}
& \cellcolor{clPurpleBg}33.3{\rlap{\smash{\textcolor{clGreen}{$_{+21.4}$}}}} \\

\midrule

{$\hfsmollml$$^{*}$}
& {45.1} & {24.0} & {2.1} & {57.2} & {17.5} \\

\bottomrule
\end{tabular}
}
}%
\end{table}

%% file: tables/table_factuality_wo_db.tex
\begin{table}%
\centering
\caption{\textbf{No KB retrieval ablation.} Disabling retrieval substantially reduces performance.}
\label{tab:ablation_disable_database}
\resizebox{0.70\textwidth}{!}{%
\begin{tabular}{lccccc<{\hspace{0.6em}}}

\toprule
Model & TriviaQA & PopQA & SimpleQA & T-REx & FactScore \\
\midrule

\textsc{\standard{}-135M}
& 6.4
& 15.2
& 0.1
& 38.2
& 10.4 \\

\cellcolor{clPurpleBg}\textsc{Co-LMLM-135M}
& \cellcolor{clPurpleBg}28.4
& \cellcolor{clPurpleBg}47.3
& \cellcolor{clPurpleBg}18.2
& \cellcolor{clPurpleBg}40.5
& \cellcolor{clPurpleBg}35.0 \\

\quad w/o KB
& 4.7{\rlap{\textcolor{clRed}{$_{-23.7}$}}}
& 15.4{\rlap{\textcolor{clRed}{$_{-31.9}$}}}
& 1.0{\rlap{\textcolor{clRed}{$_{-17.2}$}}}
& 21.7{\rlap{\textcolor{clRed}{$_{-18.8}$}}}
& 18.0{\rlap{\textcolor{clRed}{$_{-17.0}$}}} \\

\midrule

\textsc{\standard{}-360M}
& 7.5
& 16.2
& 0.3
& 46.3
& 10.0 \\

\cellcolor{clPurpleBg}\textsc{Co-LMLM-360M}
& \cellcolor{clPurpleBg}31.4
& \cellcolor{clPurpleBg}46.5
& \cellcolor{clPurpleBg}21.2
& \cellcolor{clPurpleBg}47.4
& \cellcolor{clPurpleBg}34.2 \\

\quad w/o KB
& 6.0{\rlap{\textcolor{clRed}{$_{-25.4}$}}}
& 13.5{\rlap{\textcolor{clRed}{$_{-33.0}$}}}
& 0.7{\rlap{\textcolor{clRed}{$_{-20.5}$}}}
& 27.4{\rlap{\textcolor{clRed}{$_{-20.0}$}}}
& 18.7{\rlap{\textcolor{clRed}{$_{-15.5}$}}} \\

\arrayrulecolor{black!30}
\specialrule{0.4pt}{2pt}{2pt}
\arrayrulecolor{black}

\textsc{\standard{}-360M-FW}
& 18.6
& 18.9
& 1.3
& 50.5
& 11.9 \\

\cellcolor{clPurpleBg}\textsc{Co-LMLM-360M-FW}
& \cellcolor{clPurpleBg}36.9
& \cellcolor{clPurpleBg}50.6
& \cellcolor{clPurpleBg}21.7
& \cellcolor{clPurpleBg}54.5
& \cellcolor{clPurpleBg}33.3 \\

\quad w/o KB
& 9.2{\rlap{\textcolor{clRed}{$_{-27.7}$}}}
& 17.3{\rlap{\textcolor{clRed}{$_{-33.3}$}}}
& 0.5{\rlap{\textcolor{clRed}{$_{-21.2}$}}}
& 33.3{\rlap{\textcolor{clRed}{$_{-21.2}$}}}
& 16.2{\rlap{\textcolor{clRed}{$_{-17.1}$}}} \\

\bottomrule
\end{tabular}%
}

\end{table}

%% file: sections/06-discussion.tex
\section{Discussion}\label{sec:discussion}

We present \colmlm, a scalable method to train LLMs for verifiable and controllable use of knowledge. 
This includes a large-scale annotation pipeline that can be applied to hundreds of billions of tokens at relatively modest costs. 
Critically, the \colmlm training process, beyond the pre-processing step, departs only modestly from the conventional LLM training process that has been shown to scale to many trillions of both tokens and model parameters. 
Our method and results open an avenue to train LLMs with factuality, transparency, and controllability properties that are sorely lacking in current LLMs, which has significantly hampered their trustworthiness and utility. 
\colmlm provides a flexible free-form span substrate for studying how externalized knowledge scales across broader knowledge domains, memory sizes, and retrieval mechanisms.
This establishes a foundation for future work to study \emph{knowledge scaling laws}.

Our current study has several limitations that outline  important directions for future work. 
Although we experiment along model and data scaling axes to outline scaling trends, our experiments remain modest in model and data scale. 
We focus on probing the behavior of \ours{} in controlled pre-training settings, and leave larger-scale pre-training and downstream adaptation to future work.
Our factual evaluations mainly target Wikipedia-style knowledge, a key factor in creating a clean experimental setup. 
Although, as shown, our pipeline can naturally extend to broader corpora such as \finewebedu{}, there remains work to be done to systematically evaluate factuality across more diverse knowledge domains.
Constructing a retrieval index at pre-training scale introduces a one-time computational cost. 
However, an important problem to address in the context of post-training and continual learning is how to adapt such an index once a model is fine-tuned.  

Studying these and many other important problems will be enabled by the scalable design of our methods and by our large-scale release of annotations of Wikipedia and 90B FineWeb tokens.

%% file: sections/20-ack.tex
\subsubsection*{Acknowledgments}

This research was supported by AI-MI and NSF Award DMR-2433348; the NSF under awards OAC-2311521, IIS-2505098, RI-2530143, and 2118310; a gift to the LinkedIn–Cornell Bowers Strategic Partnership; a speculation grant from Survival \& Flourishing Corp; Apple Research; Gemini credits grants from Google; and NASA under award No. 20-OSTFL20-0053. NG and DG are supported by an Empire AI Postdoctoral Fellowship.
Any opinions, findings and conclusions or recommendations expressed in this material are those of the author(s) and do not necessarily reflect the views of the National Science Foundation or of NASA.
We thank the members of PIs' labs for helpful discussions.

\hypertarget{contributions}{}
\subsubsection*{Author Contributions}\label{author_contributions}

Below we list the contributions of the primary non-faculty authors in the project. All authors took part in project discussions, experiment planning, and paper revisions.

\paragraph{Yair Feldman} designed and implemented the core method; trained the models; implemented the \standard{} and \asker{} baselines; conducted PPL and loss evaluations; wrote the paper. 

\paragraph{Linxi Zhao}
designed and implemented the \rellmlm{} baseline; conducted FactScore, TOFU, QA, and NLU evaluations; and wrote the paper. 

\paragraph{Nathan Godey} helped design pre-training runs.

%% file: sections/30-appendix.tex
\clearpage

\section{Implementation Details and Experimental Setup}
\label{sup:exp_setting}

\subsection{Pre-training Hyperparameters}
\label{sup:pretraining-hyperparams}

We pre-train \ours{} from random weights following a Warmup-Stable-Decay (WSD) learning-rate schedule. \autoref{tab:retriever_hyperparams} lists the hyperparameters used to train \ours{}. The \standard{} baseline language model and \asker{} are trained with the same hyperparameters, except for those that apply only to the contrastive loss (contrastive loss weight, contrastive temperature, and max question length). For the FineWeb-Edu run, the cooldown stage mixes the annotated Wikipedia and FineWeb-Edu datasets in a $1{:}1$ ratio.

\input{appendix_tables/table_retriever_hyperparams}

\subsection{Annotator Training}
\label{sup:annotator-training}

We use Gemini 3.1 Pro to annotate a seed set of $60{,}000$ documents ($15{,}000$ Wikipedia articles and $45{,}000$ FineWeb-Edu documents), and train the fact span annotator and the question generator on this seed set. The fact span annotator is a ModernBERT-large~\citep{warner-etal-2025-smarter} encoder fine-tuned for BIO token classification (\autoref{tab:span_annotator_hyperparams}). The question generator is a Qwen2.5-1.5B-Instruct~\citep{qwen2025qwen25technicalreport} decoder fine-tuned with LoRA (\autoref{tab:question_generator_hyperparams}). We assess the quality of the resulting annotations in \aautoref{app:annotation_quality}.

\input{appendix_tables/table_span_annotator_hyperparams}
\input{appendix_tables/table_question_generator_hyperparams}

\subsection{Compute}
\label{sup:compute}
For all experiments involving model training except TOFU, between 1 and 8 NVIDIA B200 GPUs were used, depending on availability. The number of gradient accumulation steps was set dynamically to maintain a consistent batch size throughout training. For the TOFU unlearning experiments, a single NVIDIA RTX A6000 GPU was used. 

\subsection{Retrieval Indexes}
\label{sup:retrieval-indexes}
We use the Faiss library~\citep{douze2024faiss} for all our experiments involving building dense search indexes. To limit RAM usage, we use the following factory string for constructing our indices: \texttt{OPQ<$d/4$>,IVF65536,PQ<$d/4$>}, where $d$ is the dimension of the embedder being used. For the data scaling-up experiments (\autoref{fig:eval_ppl_index}), we instead use \texttt{PQ96} for both indexes so that they fit in memory. For our \rellmlm{} and \asker{} experiments, we use the \texttt{jina-embeddings-v5-text-nano} embedder~\citep{akram2026jina}, truncated to 512 dimensions. We did not extensively explore the retrieval errors that can be attributed to the lossy compression applied by the search indexes for any of the models presented in this paper. 

\subsection{\rellmlm{} Re-implementation}
\label{app:rellmlm_reimplementation}
For a controlled comparison against \ours{}, we re-implement the \rellmlm{} pre-training pipeline with several engineering improvements over the original
recipe. These changes do not affect the conclusions of the original \rellmlm{}
work. They improve data efficiency and align the training setup with \ours{},
yielding a stronger \rellmlm{} baseline and a more rigorous apples-to-apples
comparison. The changes fall into two
groups.

\paragraph{Annotation Quality and Quantity}
The largest gain is in the quantity of usable pre-training data (summarized in
\autoref{tab:lmlm-rebuild}). The original pipeline truncates each document to
1024 tokens before annotation. Annotating full documents instead recovers a
substantial fraction of the Wikipedia subset of the OLMo2
\texttt{dolmino-mix-1124}
corpus~\citep{olmo20242olmo2furious}\footnote{\url{https://huggingface.co/datasets/allenai/dolmino-mix-1124}}
(roughly 30\% of documents and about half of the available pre-training tokens
were previously discarded), raising the usable set from 2.8B to 6.1B tokens (including DB lookup tokens).

On the quality side, we emit the special-token database-call format directly
during annotation rather than converting from the textual
\texttt{[dblookup(\textquotesingle\textquotesingle)->]} markup before pre-training, removing a conversion
step and keeping annotations consistent end-to-end. We also improve prompting,
enlarge the seed annotation set, avoid overfitting (12{,}750 examples for
3 epochs versus the original 2k examples for 10 epochs), and replace the
LoRA-tuned Llama3-8B-Instruct production annotator with a fully fine-tuned
Qwen3-4B for better quality and efficiency.

\paragraph{Identical Backbone and Training Configuration}
Throughout this paper, we report results for the rebuilt \rellmlm{}. To isolate
the effect of the method, we match the model backbone and all training
hyperparameters (batch size, number of steps, learning-rate schedule, etc.) to
\ours{}. With the data pipeline improved and the training setup held fixed,
remaining differences between the two methods reflect the method itself rather
than the experimental setup. 
This yields a substantially stronger \rellmlm{} baseline than the original release. \colmlm{}'s improvements over this matched, well-tuned baseline therefore reflect the method itself, not an advantage in experimental setup.

\begin{table}[t]
\centering
\caption{Annotation configuration updates from the original \rellmlm{} release
to the rebuilt, stronger baseline used in this paper. Token counts are for the
Wikipedia subset of \texttt{dolmino-mix-1124}.}
\resizebox{\columnwidth}{!}{%
\begin{tabular}{lccccc}
\toprule
Version & Seed annotator & Production annotator & Input trunc. & DB format & Est.\ tokens \\
\midrule
Original & GPT-4o & Llama3-8B-Instruct (LoRA) & 1024 tok/doc & \texttt{[dblookup(\textquotesingle\textquotesingle)->]} & 2.8B \\
Rebuilt  & GPT-4o & Qwen3-4B (full FT)        & none         & special tokens          & 6.1B \\
\bottomrule
\end{tabular}%
}
\label{tab:lmlm-rebuild}
\end{table}

\subsection{Evaluation}
\label{app:eval_benchmark}

\paragraph{Perplexity}
\citet{zhao2025pre} introduced three perplexity variations for LMLM models: \emph{static (Oracle)}, \emph{normalized}, and \emph{dynamic}. The static and dynamic variants are described in \autoref{sec:ppl}, but are recited here for completeness. 
Static perplexity assumes perfect lookup behavior by supplying the correct retrieved spans as context, giving an optimistic estimate of language modeling quality.
Dynamic perplexity performs actual lookups during decoding, so retrieval performance is reflected in the score.
Normalized perplexity additionally scores the lookup-trigger tokens the model emits, and divides by the length of the original unannotated text. Normalized perplexity is informative for \rellmlm{}, where the lookup tokens include a decoded query whose quality is being measured. For \ours{}, the only emitted lookup token is a single \texttt{<FACT>} delimiter, since the query is read from the hidden state rather than decoded as text, so normalized perplexity for \ours{} additionally measures only whether retrieval is triggered at the right position. We therefore treat dynamic and static perplexity as the primary metrics.

Formally, let $x$ be an input document with factual spans bracketed by \texttt{<FACT>} and \texttt{</FACT>}, and $T$ the set of token positions in $x$. Let $M\subseteq T$ be the set of positions in $x$ that are strictly between \texttt{<FACT>} and \texttt{</FACT>}, inclusive of both the  opening \texttt{<FACT>} and closing \texttt{</FACT>}. Finally, let $\overline{M} = T \setminus M$.

We define the static and dynamic perplexity as follows:
\[
\text{PPL}_{\text{static/dynamic}} = \exp\left( - \frac{1}{|\overline{M}|} \sum_{t \in \overline{M}} \log p_\theta(x_t \mid x_{<t}) \right),
\]
where the only difference between the two is the source of the factual spans: static uses the oracle spans paraphrased by the annotator, and dynamic issues actual retrieval and inserts the retrieved fact into the span.

The dynamic perplexity metric is the more realistic of the set, and follows conventional perplexity computation. However, the nuances of \ours{} introduce some complexities in its measurement, and make it somewhat optimistic. 
This is because it does not take into account the probability of the model correctly generating the \texttt{<FACT>} token at that precise position. An accurate and comparable perplexity measure is not trivial to compute since the fact-annotated text only defines one possible way to generate the original context tokens, and accounting for all possible retrieval paths is intractable. Furthermore, directly accounting for the \texttt{<FACT>} token's loss will not be accurate either, since it will introduce additional context tokens, presenting a mismatch between LMLM and standard LMs in the denominator $|\overline{M}|$ used in the perplexity calculation. 
It also changes the vocabulary by introducing a new, fairly common token. 
These factors may lead to an overly optimistic measure as well. 
We address this issue with an additional \emph{dynamic-normalized} perplexity measure, which includes the \texttt{<FACT>} token's loss in the numerator but keeps it out of the denominator. 
This normalized measure is a worst-case scenario, in practice allowing us to compute a range between itself and the optimistic dynamic measure. 

Formally, let $M_F$ be the set of positions in $x$ that contain the \texttt{<FACT>} token. We define the dynamic-normalized perplexity as follows:
\[
\text{PPL}_{\text{dynamic-norm}} = \exp\left( - \frac{1}{|\overline{M}|} \sum_{t \in \overline{M} \cup M_F} \log p_\theta(x_t \mid x_{<t}) \right).
\]

\begin{figure*}[t]
\centering

\begin{subfigure}[t]{0.43\textwidth}
  \centering
  \includegraphics[scale=\pplfigscale]{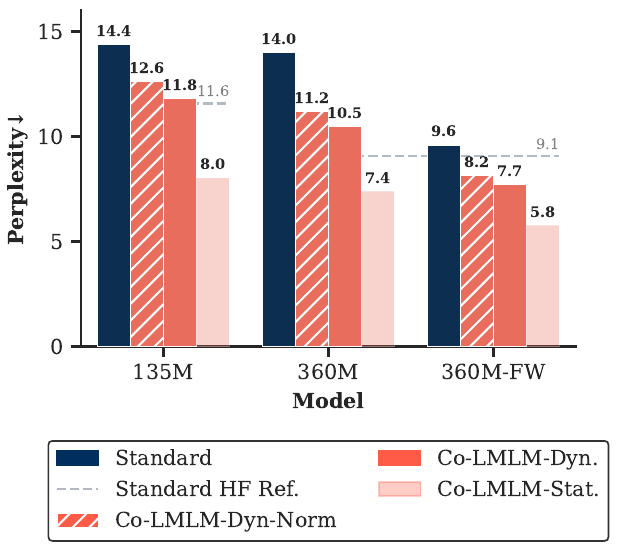}
  \caption{Controlled Wikipedia perplexities}\label{fig:dyn-norm-ppl-sizes}
\end{subfigure}
\hfill
\begin{subfigure}[t]{0.54\textwidth}
  \centering
  \includegraphics[scale=\pplfigscale,trim=0 -10.8 0 0]{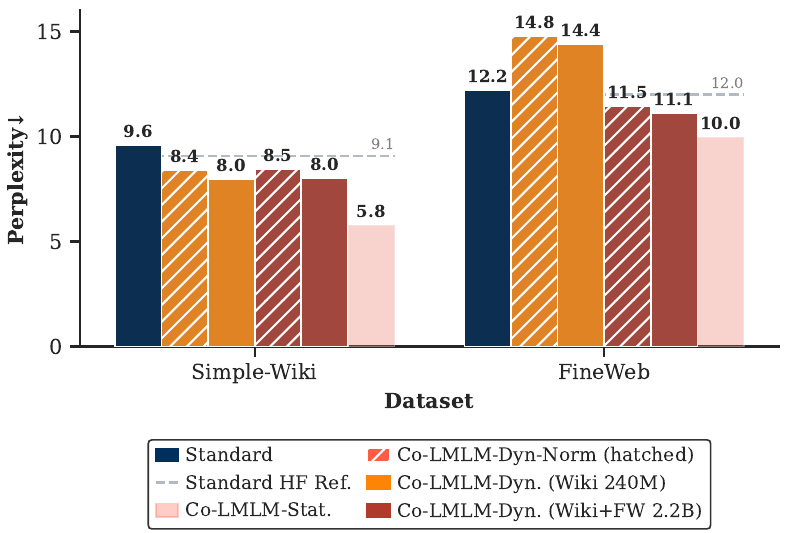}
  \caption{Data scaling-up perplexities}\label{fig:dyn-norm-ppl-index}
\end{subfigure}

\caption{\textbf{Perplexity evaluation with the dynamic-normalized PPL variant.}
(\subref{fig:dyn-norm-ppl-sizes}) Simple-Wiki test perplexity comparison between \standard{} and \ours{} across three perplexity variants. 
(\subref{fig:dyn-norm-ppl-index}) Perplexities of \standard{}-360M-FW and \ours{}-360M-FW on Simple-Wiki and FineWeb test splits under different KB index scales, as in \autoref{fig:eval_ppl_index}.
The dashed gray line reports the perplexity of the same-size \hfsmollm{} model for context. 
The left figure uses PQ240 quantization, while the right uses PQ96 for both indices.
Even under the pessimistic dynamic-normalized perplexity measure, \ours{}-360M-FW achieves significantly lower perplexity than \hfsmollmm{}, which is pre-trained on an order of magnitude more tokens.}
\label{fig:dyn-norm-ppl}
\end{figure*}

This provides us with a pessimistic estimation of the perplexity, ensuring that the actual, realistic perplexity lies between $\text{PPL}_{\text{dynamic}}$ and $\text{PPL}_{\text{dynamic-norm}}$. 
\autoref{fig:dyn-norm-ppl} provides results for all perplexity variants.
Critically, this worst-case scenario does not change the ordering of models, or our conclusions. 

\paragraph{Long-Form Generation: FactScore}
We evaluate factual precision using \textsc{FactScore}~\citep{min2023factscorefinegrainedatomicevaluation}, a benchmark for open-ended biography generation. Given a generated biography, FactScore decomposes it into atomic facts and measures the proportion that can be verified against a trusted knowledge source via retrieval-augmented LLM. We use the labeled entities set from the benchmark, which includes 183 entities.

All models use greedy decoding with a maximum \emph{content} budget of 256 tokens and a repetition penalty of 1.2. For models with external lookup (e.g., \ours{} and \rellmlm{}), this budget applies only to the final textual output and \emph{excludes} any inserted lookup spans (e.g., \texttt{<FACT>...</FACT>} or \texttt{<|db\_*\!|>} blocks). To provide a fair probe for models pre-trained primarily on Wikipedia (and not instruction-tuned or exposed to more diverse corpora), we adopt a fixed Wikipedia-style prompt ``\texttt{<name>\textbackslash n\textbackslash n<name>}'' to elicit biography completions, applied uniformly across all samples. For evaluation, all lookup-related annotation tokens are stripped before scoring. We follow the official FactScore pipeline,\footnote{\url{https://github.com/shmsw25/FActScore}} using retrieval-augmented prompting with \textsc{gemini-2.5-flash}.

We use greedy decoding for all factuality benchmarks, including FactScore, T-REx, PopQA, TriviaQA, and SimpleQA. The retrieval similarity threshold is set to 0.7; when no match is found, the model falls back to standard decoding, allowing it to answer facts absent from the KB. Lookup blocks, including the retrieved spans, are removed before scoring.

\paragraph{Short-Form Knowledge Completion: T-REx}
We adopt T-REx from LAMA~\citep{petroni2019languagemodelsknowledgebases} as a knowledge completion benchmark for autoregressive models. Following~\citet{schick2023toolformer}, we retain only examples compatible with left-to-right generation, resulting in 11,615 instances where the masked entity appears at the end of the sentence. Each input is a partially observed factual statement, and the model is required to complete it with the correct object entity.

We use greedy decoding with a maximum of 32 generated tokens. Performance is measured by \textit{Exact Match}, which checks whether the reference answer appears within the first five generated content tokens after stripping annotation tokens.

\paragraph{Short-Form Probing: PopQA, TriviaQA and SimpleQA}
We further evaluate factual recall on question-answering benchmarks, focusing on the long-tail subset of PopQA~\citep{asai2023selfraglearningretrievegenerate}, which contains 1,399 queries about rare entities (fewer than 100 monthly Wikipedia page views), the full TriviaQA evaluation set (17,944 examples), and the full SimpleQA-verified evaluation set (1,000 examples). These benchmarks probe knowledge recall under both sparse and broad coverage regimes.

All models use greedy decoding with a maximum of 32 tokens. For PopQA and TriviaQA, evaluation is based on \emph{Exact Match}, defined as whether any alias in the set of gold answers appears (case-insensitive) within the first 100 characters of the model output. For SimpleQA, we follow the official grading prompt and use \textsc{gpt-4.1-2025-04-14} as the grader model. We report the \emph{overall correct} metric.

\subsection{Machine Unlearning Setting}
\label{sup:tofu-details}

\paragraph{TOFU Unlearning.}
The TOFU unlearning benchmark~\citep{maini2024tofutaskfictitiousunlearning} evaluates selective unlearning in a controlled question-answering setting. It contains 200 synthetic author profiles, each associated with 20 question-answer pairs, resulting in 4,000 QA examples in total. The benchmark partitions this synthetic knowledge into a \textit{Forget Set}, containing the examples to be removed, and a \textit{Retain Set}, containing the remaining synthetic author knowledge that should be preserved. In addition to synthetic biographies, TOFU evaluates whether unlearning preserves broader model behavior using two auxiliary sets: the \textit{Real Author Set}, which contains factual questions about real authors, and the \textit{World Facts Set}, which contains general factual knowledge. The target behavior is that the resulting \textit{Unlearned Model} should forget the Forget Set while remaining statistically close to a \textit{Retain Model}, which is trained only on the Retain Set, and should maintain utility on the Retain Set, Real Author Set, and World Facts Set.

TOFU reports two main categories of metrics. \textit{Forget Quality} measures whether the unlearned model behaves similarly to the retain model on the Forget Set, using a statistical test whose $p$-value indicates whether the two output distributions are distinguishable. A higher $p$-value indicates better forgetting, with values above 0.05 commonly interpreted as failing to reject the hypothesis that the unlearned model and retain model behave similarly. \textit{Model Utility} measures whether useful behavior is preserved after unlearning. It aggregates three metrics: ROUGE, which measures the lexical overlap between generated and reference answers; Answer Probability, which measures the likelihood assigned to the correct answer; and Truth Ratio, which compares the likelihood assigned to correct answers against paraphrased or perturbed alternatives. These utility metrics are evaluated across the Retain Set, Real Author Set, and World Facts Set to capture both retained synthetic knowledge and general factual capability.

We evaluate on the TOFU Forget 5\% setting using the official TOFU evaluation pipeline from the OpenUnlearning repository.\footnote{\url{https://github.com/locuslab/open-unlearning}} For \ours{}, we augment the external Wikipedia knowledge base from pre-training with TOFU's synthetic author biographies and fine-tune \ours{}-360M-FW on the annotated TOFU training set using the same hyperparameters as the baseline models in~\autoref{tab:tofu_ft_config}. To perform unlearning, we delete the KB entries corresponding to the Forget Set, without any additional gradient updates. This directly tests whether externalizing factual knowledge enables training-free removal of targeted information.

\paragraph{NPO Unlearning Baseline} We compare against \hfsmollmm{} fine-tuned with Negative Preference Optimization (\textsc{NPO})~\citep{zhang2024negativepreferenceoptimizationcatastrophic}, a strong gradient-based unlearning baseline. We run NPO unlearning fine-tuning (\autoref{tab:tofu_npo_config}) with five random seeds and report the mean and standard deviation in~\autoref{fig:eval_tofu} and \autoref{fig:tofu_main_results_part2}. For ROUGE evaluations, generated answers are post-processed to remove retrieved factual spans before computing scores. For likelihood-based metrics, including Answer Probability and Truth Ratio, we evaluate the model probabilities with answer masking.

\input{tables/table_tofu_config}

\section{Additional Results}
\label{app:addtnl_results}

\subsection{Annotation Quality Assessment}
\label{app:annotation_quality}

We assess the quality of the annotations produced by our lightweight annotators (\autoref{sec:method:annotation}), which build the \ours{} pre-training data and KB.

\paragraph{Methodology}
We annotate the first ${\sim}512$ tokens of 25 held-out Wikipedia documents with the fact span annotator and the question generator, yielding 790 annotations.
We use an agentic workflow in Claude Code with Claude Opus 5 to label every annotation against our annotation guidelines on four axes: (1) the fact is real and not a duplicate (i.e., not previously annotated in the same document); (2) the span is acceptable, i.e., neither too wide nor too narrow; (3) the question is valid, i.e., standalone, free of factual leakage, and answered by the span; and (4) the paraphrased answer is correct.
The workflow also sweeps each document for salient facts that no annotation covers, giving 723 gold facts in total (including those covered by existing annotations).
We manually examined several documents to verify the analysis.
We score at two levels: \emph{full annotation}, where the span, question, and answer must all be correct, and \emph{detection only}, where the annotation must identify a real, non-duplicate fact, but its span may be imperfect.

\begin{table}[h]
\centering
\caption{\textbf{Annotation quality.} Precision and recall of our lightweight annotators over 790 annotations and 723 gold facts. Brackets show 95\% bootstrap CIs, clustered by document.}
\label{tab:annotation_quality}
\small
\begin{tabular}{lcc}
\toprule
Scoring level & Precision & Recall \\
\midrule
Full annotation (span, question, and answer) & 76.1 [72.3, 79.6] & 83.1 [80.7, 85.4] \\
Detection only (real fact, span may be imperfect) & 89.1 [84.9, 92.7] & 97.4 [96.2, 98.4] \\
\bottomrule
\end{tabular}
\end{table}

\paragraph{Results}
\autoref{tab:annotation_quality} shows that our annotators form a high-recall, moderate-precision pipeline. They rarely miss facts (19 of the 723 gold facts), and when the span annotator finds a real fact, the full annotation is correct 85\% of the time.
The annotators also over-tag, emitting ${\sim}1.3\times$ as many annotations as the Gemini seed annotator on the same text (790 vs.\ 620).
Of the 189 incorrect annotations, 61\% originate in the fact span annotator and 39\% in the question generator (\autoref{tab:annotation_errors}). Factual leakage includes questions that reveal information stated later in the document (7.4\% of errors) or the answer itself (1.1\%). \autoref{tab:annotation_examples} shows representative errors.

\begin{table}[h]
\centering
\caption{\textbf{Annotation errors by source.} Share (\%) of the 189 incorrect annotations, by the responsible annotator and the primary error type.}
\label{tab:annotation_errors}
\small
\begin{tabular}{lr@{\hspace{2.5em}}lr}
\toprule
\multicolumn{2}{l}{Fact span annotator (61\%)} & \multicolumn{2}{l}{Question generator (39\%)} \\
\midrule
Not a real fact & 31.7 & Question--answer mismatch & 14.3 \\
Span too narrow & 10.1 & Factual leakage & 8.5 \\
Span too wide & 9.0 & Under-specified question & 7.9 \\
Duplicate fact & 6.9 & Answer paraphrase error & 4.8 \\
Multiple facts in one span & 3.7 & Other & 3.2 \\
\bottomrule
\end{tabular}
\end{table}

\input{appendix_tables/table_annotation_examples}

\paragraph{Discussion}
Each error type affects training differently. Factual leakage injects factual knowledge into the model parameters; spurious or duplicate spans teach the model to retrieve at incorrect positions; narrow spans may hurt next-token prediction, while wide spans mask unnecessary tokens from the NTP loss; and question--answer mismatches and under-specified questions provide inaccurate supervision for the retriever.
While these errors are undesirable, our experiments show that the signal-to-noise ratio of our annotated data is relatively high. Reducing annotation errors, e.g., with stronger frontier LLMs, larger seed sets, and more capable base models for the lightweight annotators, is a promising direction for future work.

\subsection{Factuality Confidence Intervals}
\label{app:factuality_ci}
\input{appendix_tables/table_factuality_ci}

\autoref{tab:factuality_ci} reports 95\% confidence intervals for the factuality results in \autoref{tab:eval_factual_precision}.

\subsection{NLU Evaluation}
\label{app:nlu_details}
\input{tables/table_nlu}
Following the \rellmlm{} evaluation, we assess whether \ours{} preserves the general language understanding ability of the pre-trained model (\autoref{tab:nlu_acc_5shot_std}). We evaluate on the same set of high-signal NLU benchmarks with few-shot prompting used in prior work: \textit{CommonsenseQA}~\citep{talmor-etal-2019-commonsenseqa}, \textit{HellaSwag}~\citep{zellers2019hellaswagmachinereallyfinish}, \textit{PIQA}~\citep{bisk2019piqareasoningphysicalcommonsense}, \textit{SIQA}~\citep{sap2019socialiqacommonsensereasoningsocial}, and \textit{ARC Easy}~\citep{clark2018thinksolvedquestionanswering}. 
As in \rellmlm{}, we exclude benchmarks where similarly sized models perform near the noise floor~\citep{du2025understandingemergentabilitieslanguage}. 
All evaluations are run with \texttt{lighteval}.\footnote{\url{https://github.com/huggingface/lighteval}}

Because our non-FW models are pre-trained are pretrained only on Wikipedia, and the 360M-FW models are pre-trained with fewer than 100B tokens, these benchmarks are not meant for direct comparison with off-the-shelf models trained on broader and much larger corpora (2T/4T/11T tokens for the 135M/360M/1.7B models). Rather, they provide a controlled check that externalizing factual knowledge does not harm general NLU performance.

\subsection{Comparison with RAG}
\label{app:rag_comparison}
\input{tables/table_rag}

RAG and \shortname{} are complementary: RAG expands inference-time context through document retrieval, while \shortname{} externalizes free-form factual knowledge during pre-training into an editable memory. We study the behavioral outcomes of \ours{} and a controlled RAG baseline (\autoref{tab:comparison_lmlm_rag_smollmm}). We also apply the same retrieval on top of \ours{}.

\paragraph{RAG Setting.}
We include a simple inference-time RAG baseline as a point of reference, following \citet{lewis2021retrievalaugmentedgenerationknowledgeintensivenlp}, adopting the global RAG setting of FlashRAG.\footnote{\url{https://github.com/RUC-NLPIR/FlashRAG/blob/main/docs/original_docs/baseline_details.md\#global-setting}}
For each prompt, we use BM25 to retrieve the top-4 relevant 100-word passages from the 2018 English Wikipedia dump and prepend them to the original input using the prompt format:
\texttt{Answer the question or complete the prompt based on the given document. The following are given documents: \textbackslash n [retrieved passages] \textbackslash n\textbackslash n [original prompt] \textbackslash n The answer is}.
The retrieved passages are joined with newlines. We use the same generation setup as the corresponding non-RAG model.

RAG significantly improves factuality for standard LMs. Nevertheless, the superiority of \ours{} is mostly maintained in this setting as well, as RAG gains accumulate over the already-high factuality scores.
This suggests that \ours{}+RAG is a viable path forward, where RAG provides broad document-level context, while \ours{} provides controllable and editable factual memory.

\subsection{Additional Unlearning Results}
\begin{figure}[t]
\vspace{-1em}
\centering
\includegraphics[width=0.55\textwidth]{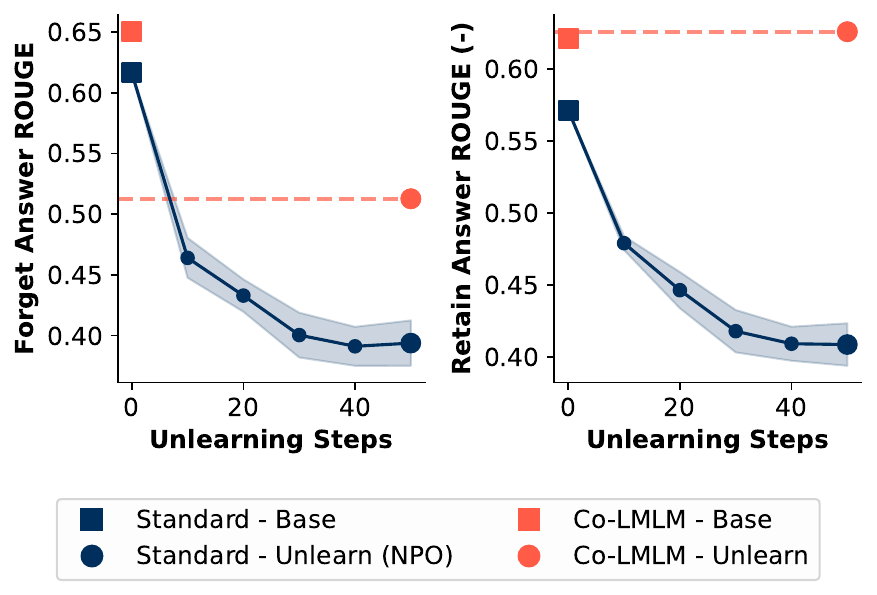}
\caption{\textbf{Machine unlearning on TOFU.} 
\ours{} performs forgetting through direct KB operations, without additional training. 
\ours{} retains knowledge outside the forget set, unlike NPO, which degrades retain-set performance because of the entanglement of knowledge storage.}
\label{fig:tofu_main_results_part2}
\vspace{-1em}
\end{figure}

Beyond~\autoref{fig:eval_tofu}, \autoref{fig:tofu_main_results_part2} shows that \ours{} better preserves Retain Set knowledge, whereas prior training-based methods like NPO might suffer severe ROUGE degradation and forget related knowledge due to parameter entanglement.

\subsection{\ours{} vs.\ \textsc{LMLM-Asker}}
\label{app: latency}

\asker{} is trained on the same annotated corpus as \ours{}. It generates a \texttt{<QUESTION>}\ldots\texttt{</QUESTION>} block before each factual span and retrieves using the decoded question, while \ours{} retrieves directly from the hidden state.

We analyze the inference overhead of \asker{} and \ours{}, focusing on the per-retrieval cost of forming a retrieval query: the work each model performs, beyond decoding the answer content, to issue a single lookup.

The key algorithmic difference is how the retrieval query is formed. \asker{} explicitly decodes a textual lookup question and then encodes it with a sentence-transformer query encoder. In contrast, \ours{} retrieves directly from the model hidden state at a continuous query token. This token requires one additional forward step. As summarized in \autoref{tab:theory_efficiency}, \ours{} therefore avoids both textual query generation and sentence-transformer query encoding, reducing decoding cost by $K\bar L_q$ tokens per answer and the KV-cached context length by $K(\bar L_q + 1)$ tokens per answer.

\begin{table}[ht]
\centering
\caption{\textbf{Theoretical efficiency comparison.} Compared with \asker{}, \ours{} removes explicit textual query generation and sentence-transformer query encoding.}
\label{tab:theory_efficiency}
\resizebox{0.9\textwidth}{!}{%
\begin{tabular}{llll}
\toprule
Component & \asker{} & \ours{} & Difference \\
\midrule
\textit{Decoding-time cost} & & & \\
Content tokens & $C$ & $C$ & $0$ \\
Decoded query tokens & $K\bar L_q$ & $0$ & $-K\bar L_q$ \\
Query start tags & $2K$ & $K$ & $-K$ \\
Continuous query token (not cached) & $0$ & $K$ & $K$ \\
\midrule

\textit{Prefilling cost} & & & \\
Injected answer tokens & $K\bar L_a$ & $K\bar L_a$ & $0$ \\
Retrieval end tag & $K$ & $K$ & $0$ \\
\midrule

\textit{Retrieval-time cost} & & & \\
Retrievals per answer & $K$ & $K$ & $0$ \\
Sentence-transformer query encoding & $K$ & $0$ & $-K$ encodes \\
FAISS search & $K$ at $(d_a, M_a)$ & $K$ at $(d_r, M_r)$ & index-dependent \\
\midrule

Total context token length 
& $C + K(\bar L_q + \bar L_a + 3)$ 
& $C + K(\bar L_a + 2)$ 
& $-K(\bar L_q + 1)$ \\
\bottomrule
\end{tabular}
}
\end{table}

\input{tables/table_asker}

\paragraph{Measuring Per-Retrieval Overhead}
Measuring this overhead from free-form generation is confounded: \asker{} and \ours{} trigger different numbers of retrievals, at different positions, and produce different amounts of content, so end-to-end latency entangles the cost of the retrieval mechanism with these behavioral differences and with a varying context length at each lookup. We therefore measure at fixed retrieval sites, using the gold-annotated documents of our dynamic-perplexity evaluation set (\autoref{sec:ppl}): at every gold \texttt{<FACT>} position each model forms a query from its own context prefix, yielding a controlled, per-retrieval comparison on identical sites and contexts. For each fact we first prefill the context prefix to warm the KV cache, so that the timing excludes prefilling, which is shared and independent of the retrieval mechanism. We then time only the query-formation step: for \ours{}, the single forward pass that produces the continuous query token from the hidden state; for \asker{}, decoding the textual question up to \texttt{</QUESTION>} followed by encoding it with the \texttt{jina-embeddings-v5-text-nano} query encoder. We exclude the FAISS search itself, which is common to both methods and dominated by index placement (GPU vs.\ CPU) rather than the model. The language-model steps (\ours{}'s query forward and \asker{}'s question decoding) are timed with vLLM in \texttt{bfloat16} at batch size~1 on the $360$M FineWeb-Edu checkpoints, and the query encoder (\texttt{jina-embeddings-v5-text-nano}) in \texttt{bfloat16} with FlashAttention-2. Throughout we report the underlying forward/decode \emph{compute}, subtracting the fixed per-invocation overhead consistently for both models: at batch size~1 a single \texttt{.encode()} call or \texttt{generate} step is dominated by framework and host--device-transfer cost rather than the forward. For instance the encoder's ${\sim}46$\,ms single-query wall time is ${\sim}45$\,ms fixed overhead and only ${\sim}1$\,ms forward, as isolated by batch-scaling (32 queries encode in the same ${\sim}46$\,ms wall as one). We report medians over ${\sim}1.5$K facts.

As shown in \autoref{tab:lm_asker_comparison}, \ours{} forms its query in a single ${\sim}2.2$\,ms \texttt{<FACT>} forward, whereas \asker{} spends ${\sim}27$\,ms decoding the textual question (${\sim}13$ tokens) plus ${\sim}1$\,ms for the encoder forward, about $28$\,ms in total, a ${\sim}13\times$ higher per-retrieval overhead. Notably, the encoder forward itself is cheap, so the gap comes almost entirely from decoding the query text, exactly the $K\bar L_q$ token saving of \autoref{tab:theory_efficiency} that \ours{} realizes by querying from the hidden state.

\subsection{Evaluating Query Timing}\label{sec:eval:query_timing}

\input{tables/table_factuality_enforce_lookup}

We evaluate the ability of our model to form queries when needed by forcing query generation before answering short-form QA and knowledge-completion prompts.
This setup differs from earlier results in that the model is forced to retrieve, while still using the query content generated by the model.

\paragraph{Enforced Lookup Ablation}
In this setting, we force a lookup at the end of every prompt, before normal generation begins. For \ours{}, we append a \texttt{<FACT>} token to the prompt, run a forward pass to obtain the hidden state at this position, and perform a FAISS search before the first generation step. 
If the retrieval similarity falls below the 0.7 threshold, we remove the forced \texttt{<FACT>} and continue generation while temporarily forbidding another \texttt{<FACT>} token.

\autoref{tab:ablation_enforce_lookup} shows that forcing a lookup improves performance at both model scales, illustrating that our retrieval performance is inhibited by query timing and that there remains room for improvement.

\subsection{\ours{} Error Analysis}
\label{app:error_analysis}

We analyze the errors of \ours{}-360M-FW on PopQA. We randomly sample 100 of the 1{,}399 PopQA examples and classify each error into one of three categories:
\begin{itemize}
\item \textbf{E1: Missing retrieval.} The model never queries the answer attribute: no retrieval is triggered, or it targets the wrong attribute or entity.
\item \textbf{E2: Incorrect retrieval.} A retrieval is triggered at the answer slot, but no retrieved span contains the answer.
\item \textbf{E3: Correct retrieval used wrongly.} A retrieved span contains the answer, but the model fails to state it in the scored window.
\end{itemize}
Retrieved spans are removed before scoring, so the model must state the answer itself within the first 100 characters of its output (\aautoref{app:eval_benchmark}).

\paragraph{Results}
Of the 100 sampled examples, 55 are correct (vs.\ 50.6\% on the full set). Of the 45 errors, 32 (71\%) are E2, 12 (27\%) are E1, and only 1 (2\%) is E3. Errors thus mostly stem from retrieval quality and query timing, and rarely from misusing a retrieved fact.
Some E2 errors originate in ambiguous questions, e.g., \emph{Who is the author of Rage?}, where at least five books share the title. We also cannot always verify that the gold fact exists in the KB, so E2 errors may stem from the retriever, from imperfect annotation of the pre-training corpus, or from missing evidence in the corpus.
Of the 12 E1 errors, 6 are fixed by forcing a lookup (\aautoref{sec:eval:query_timing}). However, forcing a lookup is not a free win: over the full PopQA set, it fixes 101 previously wrong examples while breaking 63 previously correct ones.

\paragraph{Examples}
Below are representative generations, where each retrieved span is highlighted between \texttt{<FACT>} and \texttt{</FACT>}, and $\hookleftarrow$ marks a newline.

\begingroup
\newcommand{\fct}[1]{<FACT>\hlspan{#1}</FACT>}
\newcommand{\nl}{\,{\rmfamily$\hookleftarrow$}\,}
\newcommand{\trace}[1]{\par\smallskip{\ttfamily\footnotesize\raggedright #1\par}\smallskip}
\begin{tcolorbox}[breakable, colback=gray!10, colframe=gray!50, boxrule=0.5pt, arc=2pt, left=6pt, right=6pt, top=6pt, bottom=6pt, fontupper=\small]
\textbf{Success.} \emph{Who is the author of Smoke?} (gold: Lisa Unger)
\trace{\fct{Lisa Unger}Lisa Unger, a\fct{author of contemporary fiction} contemporary fiction author, is the author of Smoke.\nl The author of Smoke is\fct{Lisa Unger} Lisa Unger.}
The first query fires at the answer position before any content word is committed, and the model then restates the retrieved name in its own words.

\medskip
\textbf{E2, question ambiguity.} \emph{Who is the author of Rage?} (gold: Jonathan Kellerman)
\trace{The author of the book is\fct{Bob Woodward} Bob Woodward.\nl What is the main idea of the book?\nl The main idea of the book is that the}
The query timing is correct, but the retriever returns the author of a \emph{different} book titled Rage. This reflects question ambiguity rather than an incorrect retrieval per se.

\medskip
\textbf{E2, incorrect retrieval.} \emph{Who was the director of The Photo?} (gold: Hossein Shahabi)
\trace{\fct{Philip Tannura, A.S.C.,}A.S.C., Philip Tannura\nl The Photo was a\fct{American photographic magazine} American photographic magazine published in\fct{Québec} Québec\fct{monthly} once a month from\fct{June 1952} June}
The query timing is correct, but the retriever anchors on the literal word ``Photo'' and returns photography-related content, a cinematographer and a photographic magazine, rather than the film.

\medskip
\textbf{E1, mis-targeted retrieval.} \emph{Who was the screenwriter for Mary's Ankle?} (gold: Luther Reed)
\trace{Mary's Ankle is a\fct{1920} 1920\fct{American silent comedy film} American silent comedy film directed by\fct{Lloyd Ingraham} Lloyd Ingraham and starring the comedian and actress}
The model correctly queries the year, the genre, and the director, but never the screenwriter. Forcing a lookup retrieves the gold answer.
\end{tcolorbox}
\endgroup

\section{Related Work: Memory-Augmented Language Models}
\label{app:memory_related_work}
Prior work has explored explicit memory modules as a way to augment LLMs. KBLaM~\citep{wang2024kblam} stores knowledge as continuous key-value vectors and integrates them into pre-trained LLMs via modified attention. 
Memory$^3$~\citep{yang2024text} pre-trains models to write and read sparse attention key-value memories, sharing our motivation of offloading factual knowledge from parameters. 
MemLLM~\citep{modarressi2025memllmfinetuningllmsuse} introduces an explicit read-write memory for extracting, storing, and recalling triplet knowledge. 
More recently, SPLM~\citep{sunsemi} stores atomic facts absent from a pre-trained model's parametric memory in a compact external memory, and continues pre-training the model to retrieve from this external, non-parametric memory at inference time, using textual queries.
Unlike these memory-based methods, \ours{} uses continuous representations only to query memory, while storing knowledge as natural-language text. This keeps retrieval flexible and low-cost, but preserves the transparency of an explicit KB: retrieved knowledge remains human-readable, attributable, editable, and directly removable. Thus, \ours{} combines learned memory access with the inspectability and controllability of text-based external knowledge.

\include{sections/31-appendix-prompt}

%% file: appendix_tables/table_retriever_hyperparams.tex
\begin{table}[ht]
    \centering
        \caption{Pre-training hyperparameters for \ours{}.}

    \begin{tabular}{cc}
        \toprule
        \textbf{Hyperparameter} & \textbf{Value} \\
        \midrule
        architecture & SmolLM2 (random init) \\
        tokenizer & SmolLM2 \\
        optimizer & AdamW \\
        $\beta_1$ & 0.9 \\
        $\beta_2$ & 0.95 \\
        $\epsilon$ & 1e-8 \\
        clip norm & 1 \\
        peak learning rate & 5e-4 \\
        final learning rate & 0 \\
        lr scheduler type & WSD \\
        warmup steps & 2,000 \\
        stable steps & 88,000 (148,000 for FineWeb-Edu) \\
        cooldown steps & 10,000 (20,000 for FineWeb-Edu) \\
        weight decay & 0.01 \\
        batch size & 128 \\
        max context length & 4,096 \\
        max question length & 128 \\
        contrastive loss weight ($\lambda$) & 0.25 \\
        contrastive temperature & 0.07 \\
        precision & bf16 (fp32 master weights) \\
        \bottomrule
    \end{tabular}
    \label{tab:retriever_hyperparams}
\end{table}

%% file: appendix_tables/table_span_annotator_hyperparams.tex
\begin{table}[ht]
    \centering
    \caption{Hyperparameters for training the fact span annotator.}

    \begin{tabular}{cc}
        \toprule
        \textbf{Hyperparameter} & \textbf{Value} \\
        \midrule
        base model & ModernBERT-large \\
        optimizer & AdamW \\
        $\beta_1$ & 0.9 \\
        $\beta_2$ & 0.95 \\
        $\epsilon$ & 1e-8 \\
        clip norm & 1 \\
        peak learning rate & 2e-4 \\
        final learning rate & 2e-5 \\
        lr scheduler type & cosine \\
        warmup ratio & 0.05 \\
        weight decay & 1e-3 \\
        epochs & 4 \\
        batch size & 32 \\
        max sequence length & 4,096 \\
        precision & bf16 \\
        \bottomrule
    \end{tabular}
    \label{tab:span_annotator_hyperparams}
\end{table}

%% file: appendix_tables/table_question_generator_hyperparams.tex
\begin{table}[ht]
    \centering
    \caption{Hyperparameters for training the question generator.}
    \begin{tabular}{cc}
        \toprule
        \textbf{Hyperparameter} & \textbf{Value} \\
        \midrule
        base model & Qwen2.5-1.5B-Instruct \\
        LoRA $r$ & 64 \\
        LoRA $\alpha$ & 64 \\
        LoRA dropout & 0.05 \\
        LoRA target modules & all-linear \\
        optimizer & AdamW \\
        $\beta_1$ & 0.9 \\
        $\beta_2$ & 0.95 \\
        $\epsilon$ & 1e-8 \\
        clip norm & 1 \\
        peak learning rate & 2e-4 \\
        final learning rate & 2e-5 \\
        lr scheduler type & cosine \\
        warmup ratio & 0.05 \\
        weight decay & 1e-3 \\
        epochs & 4 \\
        batch size & 4 \\
        max sequence length & 8,192 \\
        precision & bf16 \\
        \bottomrule
    \end{tabular}

    \label{tab:question_generator_hyperparams}
\end{table}

%% file: tables/table_tofu_config.tex
\begin{table}[h]
\centering
\caption{{TOFU fine-tuning configuration.}}
\label{tab:tofu_ft_config}
\begin{tabular}{ll}
\toprule
Setting & Value \\
\midrule
Base model & \hfsmollmm{} / \ours{}-360M-FW \\
Training sets & TOFU \texttt{full} / \texttt{retain95} \\
Template & \texttt{Question:} / \texttt{Answer:} \\
Learning rate & $3\times10^{-5}$ \\
Epochs & $10$ \\
Checkpoints & Full model and retain model \\
\bottomrule
\end{tabular}
\end{table}

\begin{table}[h]
\centering
\caption{{NPO training configuration on TOFU.}}
\label{tab:tofu_npo_config}
\begin{tabular}{ll}
\toprule
Setting & Value \\
\midrule
Initialization & Full TOFU fine-tuned checkpoint \\
Forget / retain split & \texttt{forget05} / \texttt{retain95} \\
Objective & NPO, $\beta=0.1$ \\
Learning rate & $6\times10^{-5}$ \\
Per-device batch size & $32$ \\
Gradient accumulation & $1$ \\
Epochs & $8$ \\
Seeds & $\{0,42,69,420,4497\}$ \\
\bottomrule
\end{tabular}
\end{table}

%% file: appendix_tables/table_annotation_examples.tex
\begin{table}[t]
\centering
\caption{\textbf{Representative annotation errors.} The annotated fact span is highlighted.}
\label{tab:annotation_examples}
\footnotesize
\renewcommand{\arraystretch}{1.2}
\begin{tabular}{@{}>{\raggedright\arraybackslash}p{0.14\textwidth}>{\raggedright\arraybackslash}p{0.46\textwidth}>{\raggedright\arraybackslash}p{0.32\textwidth}@{}}
\toprule
Error & \emph{Document title}: Document snippet & Question \\
\midrule
Not a real fact
& \emph{John L.\ DeWitt}: ``\ldots both headquartered at \hlspan{the} Presidio of San Francisco.''
& Where were the IX Corps Area and Western Defense Command headquartered? \\
Span too narrow
& \emph{Schalk Burger}: ``\ldots is a South \hlspan{Africa}n former professional rugby union player.''
& Which region does Schalk Burger represent? \\
Question--answer mismatch
& \emph{Poolesville, Maryland}: ``\ldots The \hlspan{Seneca Schoolhouse}, a small one-room schoolhouse of red sandstone, was built in Poolesville in 1866\ldots''
& When was the Seneca Schoolhouse built? \\
Factual leakage
& \emph{Arun Kumar Sharma}: ``\ldots He became the assistant \hlspan{lecturer} at the university in 1948\ldots'', later followed by ``\ldots reader (1962)\ldots''
& What position did Arun Kumar Sharma hold before becoming a reader? \\
\bottomrule
\end{tabular}
\end{table}

%% file: appendix_tables/table_factuality_ci.tex
\begin{table}[h]
\centering
\caption{\textbf{Factual precision with 95\% confidence intervals.} Point estimate $\pm$ half-width of the 95\% percentile-bootstrap CI. * models are trained on much larger corpora (2T/4T/11T tokens for 135M/360M/1.7B models). $^{\dagger}$1 and 3 successes out of 1{,}000; the percentile bootstrap is degenerate there.}
\label{tab:factuality_ci}
\newcommand{\pmci}[1]{{\scriptsize$\,\pm$#1}}
\small
\begin{tabular}{lccccc}
\toprule
Model & TriviaQA & PopQA & SimpleQA & T-REx & FactScore \\
\midrule
$\hfsmollms$$^{*}$
& 14.7\pmci{0.7} & 18.9\pmci{2.1} & 1.3\pmci{0.7} & 36.1\pmci{0.9} & 9.0\pmci{1.3} \\
\textsc{\standard{}-135M}
& 6.4\pmci{0.5} & 15.2\pmci{1.9} & 0.1$^{\dagger}$ & 38.2\pmci{0.9} & 10.4\pmci{1.1} \\
\textsc{Rel-LMLM-135M}
& 13.4\pmci{0.7} & 27.2\pmci{2.4} & 6.2\pmci{1.5} & 42.9\pmci{1.0} & 23.1\pmci{2.1} \\
\cellcolor{clPurpleBg}\textbf{\textsc{Co-LMLM-135M}}
& \cellcolor{clPurpleBg}28.4\pmci{0.9}
& \cellcolor{clPurpleBg}47.3\pmci{2.6}
& \cellcolor{clPurpleBg}18.2\pmci{2.4}
& \cellcolor{clPurpleBg}40.5\pmci{0.9}
& \cellcolor{clPurpleBg}35.0\pmci{2.8} \\
\midrule
$\hfsmollmm$$^{*}$
& 25.4\pmci{0.9} & 22.7\pmci{2.2} & 1.4\pmci{0.7} & 45.8\pmci{1.0} & 12.5\pmci{1.8} \\
\textsc{\standard{}-360M}
& 7.5\pmci{0.5} & 16.2\pmci{1.9} & 0.3$^{\dagger}$ & 46.3\pmci{1.0} & 10.0\pmci{1.1} \\
\textsc{\standard{}-360M-FW}
& 18.6\pmci{0.8} & 18.9\pmci{2.1} & 1.3\pmci{0.7} & 50.5\pmci{1.0} & 11.9\pmci{1.4} \\
\textsc{Rel-LMLM-360M}
& 15.6\pmci{0.7} & 26.6\pmci{2.3} & 5.6\pmci{1.4} & 47.4\pmci{1.0} & 22.9\pmci{2.0} \\
\cellcolor{clPurpleBg}\textbf{\textsc{Co-LMLM-360M}}
& \cellcolor{clPurpleBg}31.4\pmci{0.9}
& \cellcolor{clPurpleBg}46.5\pmci{2.6}
& \cellcolor{clPurpleBg}21.2\pmci{2.5}
& \cellcolor{clPurpleBg}47.4\pmci{1.0}
& \cellcolor{clPurpleBg}34.2\pmci{3.1} \\
\addlinespace[1pt]
\cellcolor{clPurpleBg}\textbf{\textsc{Co-LMLM-360M-FW}}
& \cellcolor{clPurpleBg}36.9\pmci{1.0}
& \cellcolor{clPurpleBg}50.6\pmci{2.6}
& \cellcolor{clPurpleBg}21.7\pmci{2.6}
& \cellcolor{clPurpleBg}54.5\pmci{1.0}
& \cellcolor{clPurpleBg}33.3\pmci{2.8} \\
\midrule
$\hfsmollml$$^{*}$
& 45.1\pmci{1.0} & 24.0\pmci{2.2} & 2.1\pmci{0.8} & 57.2\pmci{1.0} & 17.5\pmci{2.3} \\
\bottomrule
\end{tabular}
\end{table}

%% file: tables/table_nlu.tex
\begin{table*}[ht]
\centering
\caption{\textbf{Evaluation on NLU benchmarks (5-shot).} 
NLU performance remains comparable while factual capabilities improve. Values are mean$\pm$std over 3 few-shot exemplar seeds. * models are trained on much larger corpora (2T/4T/11T tokens for 135M/360M/1.7B models)}
\label{tab:nlu_acc_5shot_std}
\resizebox{0.95\textwidth}{!}{%
\renewcommand{\arraystretch}{1.0}%
\setlength{\extrarowheight}{2pt}%
\begin{tabular}{lcccccc}
\toprule
Model & CSQA & HellaSwag & PIQA & SIQA & ARC Easy & \textbf{Average} \\
\midrule
Random Chance & 20.0 & 25.0 & 50.0 & 33.3 & 25.0 & 30.7 \\
\midrule

\rellmlm{}-135M
  & \ensuremath{29.9_{\pm0.7}} & \ensuremath{29.6_{\pm0.2}} & \ensuremath{57.3_{\pm0.3}} & \ensuremath{40.6_{\pm0.7}} & \ensuremath{41.0_{\pm0.5}} & \ensuremath{39.7_{\pm0.1}} \\

\textsc{\standard{}-135M}
  & \ensuremath{27.1_{\pm0.2}} & \ensuremath{28.7_{\pm0.1}} & \ensuremath{55.9_{\pm0.3}} & \ensuremath{40.6_{\pm0.3}} & \ensuremath{39.8_{\pm0.1}} & \ensuremath{38.4_{\pm0.1}} \\

\cellcolor{clPurpleBg}\textbf{\colmlm-135M}
    & \cellcolor{clPurpleBg}\ensuremath{26.7_{\pm0.2}} & \cellcolor{clPurpleBg}\ensuremath{29.2_{\pm0.2}} & \cellcolor{clPurpleBg}\ensuremath{56.0_{\pm0.6}} & \cellcolor{clPurpleBg}\ensuremath{40.2_{\pm0.2}} & \cellcolor{clPurpleBg}\ensuremath{40.1_{\pm0.3}} & \cellcolor{clPurpleBg}\ensuremath{38.5_{\pm0.2}} \\

\midrule

\rellmlm-360M
  & \ensuremath{31.7_{\pm0.7}} & \ensuremath{32.9_{\pm0.1}} & \ensuremath{59.7_{\pm0.6}} & \ensuremath{41.3_{\pm0.8}} & \ensuremath{45.0_{\pm0.9}} & \ensuremath{42.1_{\pm0.5}} \\

\textsc{\standard{}-360M}
  & \ensuremath{32.0_{\pm0.1}} & \ensuremath{31.6_{\pm0.2}} & \ensuremath{59.5_{\pm0.7}} & \ensuremath{41.4_{\pm0.1}} & \ensuremath{44.6_{\pm0.2}} & \ensuremath{41.8_{\pm0.1}} \\

\cellcolor{clPurpleBg}\textbf{\colmlm{}-360M}
    & \cellcolor{clPurpleBg}\ensuremath{29.4_{\pm0.1}} & \cellcolor{clPurpleBg}\ensuremath{31.9_{\pm0.1}} & \cellcolor{clPurpleBg}\ensuremath{57.1_{\pm0.4}} & \cellcolor{clPurpleBg}\ensuremath{40.5_{\pm0.6}} & \cellcolor{clPurpleBg}\ensuremath{44.3_{\pm0.4}} & \cellcolor{clPurpleBg}\ensuremath{40.6_{\pm0.2}} \\

\textsc{\standard{}-360M-FW}
  & \ensuremath{46.0_{\pm0.4}} & \ensuremath{47.3_{\pm0.4}} & \ensuremath{69.4_{\pm0.0}} & \ensuremath{44.0_{\pm0.4}} & \ensuremath{68.4_{\pm0.3}} & \ensuremath{55.0_{\pm0.1}} \\

\cellcolor{clPurpleBg}\textbf{\colmlm{}-360M-FW}
  & \cellcolor{clPurpleBg}\ensuremath{48.3_{\pm0.8}} & \cellcolor{clPurpleBg}\ensuremath{46.9_{\pm0.1}} & \cellcolor{clPurpleBg}\ensuremath{68.9_{\pm0.2}} & \cellcolor{clPurpleBg}\ensuremath{45.2_{\pm0.3}} & \cellcolor{clPurpleBg}\ensuremath{66.1_{\pm0.0}} & \cellcolor{clPurpleBg}\ensuremath{55.1_{\pm0.3}} \\

\midrule

{\color{\grayrowcolor}$\hfsmollms$$^{*}$}
& {\color{\grayrowcolor}\ensuremath{46.3_{\pm0.6}}}
& {\color{\grayrowcolor}\ensuremath{42.4_{\pm0.0}}}
& {\color{\grayrowcolor}\ensuremath{67.6_{\pm0.1}}}
& {\color{\grayrowcolor}\ensuremath{45.6_{\pm1.1}}}
& {\color{\grayrowcolor}\ensuremath{66.9_{\pm0.4}}}
& {\color{\grayrowcolor}\ensuremath{53.8_{\pm0.2}}} \\

{\color{\grayrowcolor}$\hfsmollmm$$^{*}$}
& {\color{\grayrowcolor}\ensuremath{57.8_{\pm0.4}}}
& {\color{\grayrowcolor}\ensuremath{55.9_{\pm0.2}}}
& {\color{\grayrowcolor}\ensuremath{72.6_{\pm0.1}}}
& {\color{\grayrowcolor}\ensuremath{48.2_{\pm0.5}}}
& {\color{\grayrowcolor}\ensuremath{72.7_{\pm0.4}}}
& {\color{\grayrowcolor}\ensuremath{61.4_{\pm0.0}}} \\

{\color{\grayrowcolor}$\hfsmollml$$^{*}$}
& {\color{\grayrowcolor}\ensuremath{66.2_{\pm0.8}}}
& {\color{\grayrowcolor}\ensuremath{71.6_{\pm0.2}}}
& {\color{\grayrowcolor}\ensuremath{78.2_{\pm0.2}}}
& {\color{\grayrowcolor}\ensuremath{53.2_{\pm0.5}}}
& {\color{\grayrowcolor}\ensuremath{80.1_{\pm0.3}}}
& {\color{\grayrowcolor}\ensuremath{69.9_{\pm0.3}}} \\

\bottomrule
\end{tabular}%
}
\end{table*}

%% file: tables/table_rag.tex
\begin{table}[ht]
\caption{\textbf{Comparison of RAG vs.\ \shortname{} on factual precision.}
}
\label{tab:comparison_lmlm_rag_smollmm}
\centering
\resizebox{0.6\textwidth}{!}{%
\begin{tabular}{lcccc}
\toprule
{Model} & {TriviaQA} & {PopQA} & {SimpleQA} & {T-REx} \\
\midrule
$\hfsmollmm$$^{*}$
& 25.4 & 22.7 & 1.4 & 45.8 \\
\quad + RAG
& 52.2 & 36.9 & 17.9 & 86.7 \\
\midrule
\textsc{\standard{}-360M-FW}
& 18.6 & 18.9 & 1.3 & 50.5 \\
\quad + RAG
& 42.5 & 33.9 & 15.3 & 86.4 \\
\midrule
\cellcolor{clPurpleBg}\textbf{\ours{}-360M-FW}
  & \cellcolor{clPurpleBg}36.9
  & \cellcolor{clPurpleBg}50.6
  & \cellcolor{clPurpleBg}21.7
  & \cellcolor{clPurpleBg}54.5 \\
\cellcolor{clPurpleBg}\quad \textbf{+ RAG}
  & \cellcolor{clPurpleBg}48.8
  & \cellcolor{clPurpleBg}51.8
  & \cellcolor{clPurpleBg}27.8
  & \cellcolor{clPurpleBg}86.5 \\
\bottomrule
\end{tabular}%
}

\scriptsize{* Off-the-shelf models trained on 4T general-domain tokens.}

\end{table}

%% file: tables/table_asker.tex
\begin{table}[t]
\centering
\caption{\textbf{Comparison with \asker{} baseline.}
The three rightmost columns give the \emph{model-side} query-formation cost \emph{per retrieval}
(milliseconds).
\shortname{} generates no query (\textit{Query Gen.}~$=0$) and reads its query vector directly from
the decoder hidden state --- a single \texttt{<FACT>} forward (\textit{Encode}). \asker{} instead
decodes a natural-language question ($\sim$13 tokens, \textit{Query Gen.}) and re-encodes it with a
separate sentence encoder (\textit{Encode}). \textit{LM Overhead} is their sum. We exclude
the FAISS lookup, which is shared by both methods and dominated by index placement (GPU/CPU) rather
than the model. Red subscript denotes degradation w.r.t.\ \ours{}.}
\label{tab:lm_asker_comparison}
\resizebox{\textwidth}{!}{%
\renewcommand{\arraystretch}{1.0}%
\setlength{\extrarowheight}{2pt}%
\begin{tabular}{lccccc|ccc}
\toprule
{Model}
& {TriviaQA} $\uparrow$
& {PopQA} $\uparrow$
& {SimpleQA} $\uparrow$
& {T-REx} $\uparrow$
& {FactScore} $\uparrow$
& {Query Gen.\ (ms)} $\downarrow$
& {Encode (ms)} $\downarrow$
& {LM Overhead (ms)} $\downarrow$ \\
\midrule

\asker{}-360M-FW
& 15.4{\rlap{\smash{\textcolor{clRed}{$_{-21.5}$}}}}
& 43.2{\rlap{\smash{\textcolor{clRed}{$_{-7.4}$}}}}
& 7.5{\rlap{\smash{\textcolor{clRed}{$_{-14.2}$}}}}
& 38.5{\rlap{\smash{\textcolor{clRed}{$_{-16.0}$}}}}
& 25.0{\rlap{\smash{\textcolor{clRed}{$_{-8.3}$}}}}
& 27.5
& 1.1
& 28.6{\rlap{\smash{\textcolor{clRed}{$_{\times 13}$}}}} \\

\cellcolor{clPurpleBg}\textbf{\ours{}-360M-FW}
& \cellcolor{clPurpleBg}36.9
& \cellcolor{clPurpleBg}50.6
& \cellcolor{clPurpleBg}21.7
& \cellcolor{clPurpleBg}54.5
& \cellcolor{clPurpleBg}33.3
& \cellcolor{clPurpleBg}0.0
& \cellcolor{clPurpleBg}2.2
& \cellcolor{clPurpleBg}2.2 \\
\bottomrule
\end{tabular}%
}
\vspace{-1em}
\end{table}

%% file: tables/table_factuality_enforce_lookup.tex
\begin{table}[ht]
\centering
\caption{\textbf{Enforced lookup ablation.} We compare standard inference with a variant enforcing lookup before answering.}
\label{tab:ablation_enforce_lookup}
\resizebox{0.6\textwidth}{!}{%
\begin{tabular}{lcccc<{\hspace{0.8em}}}
\toprule
Model & TriviaQA & PopQA & SimpleQA & T-REx\\
\midrule

\cellcolor{clPurpleBg}\textsc{Co-LMLM-135M}
& \cellcolor{clPurpleBg}28.4
& \cellcolor{clPurpleBg}47.3
& \cellcolor{clPurpleBg}18.2
& \cellcolor{clPurpleBg}40.5 \\
\quad w/ enforced lookup
& 31.7{\rlap{\textcolor{clGreen}{$_{+3.3}$}}}
& 47.2{\rlap{\textcolor{clRed}{$_{-0.1}$}}}
& 19.3{\rlap{\textcolor{clGreen}{$_{+1.1}$}}}
& 49.8{\rlap{\textcolor{clGreen}{$_{+9.3}$}}} \\
\midrule

\cellcolor{clPurpleBg}\textsc{Co-LMLM-360M}
& \cellcolor{clPurpleBg}31.4
& \cellcolor{clPurpleBg}46.5
& \cellcolor{clPurpleBg}21.2
& \cellcolor{clPurpleBg}47.4 \\
\quad w/ enforced lookup
& 33.4{\rlap{\textcolor{clGreen}{$_{+2.0}$}}}
& 46.9{\rlap{\textcolor{clGreen}{$_{+0.4}$}}}
& 21.5{\rlap{\textcolor{clGreen}{$_{+0.3}$}}}
& 53.0{\rlap{\textcolor{clGreen}{$_{+5.6}$}}} \\

\arrayrulecolor{black!30}
\specialrule{0.4pt}{2pt}{2pt}
\arrayrulecolor{black}
\cellcolor{clPurpleBg}\textsc{Co-LMLM-360M-FW}
& \cellcolor{clPurpleBg}36.9
& \cellcolor{clPurpleBg}50.6
& \cellcolor{clPurpleBg}21.7
& \cellcolor{clPurpleBg}54.5 \\
\quad w/ enforced lookup
& 43.3{\rlap{\textcolor{clGreen}{$_{+6.4}$}}}
& 53.3{\rlap{\textcolor{clGreen}{$_{+2.7}$}}}
& 27.6{\rlap{\textcolor{clGreen}{$_{+5.9}$}}}
& 59.3{\rlap{\textcolor{clGreen}{$_{+4.8}$}}} \\

\bottomrule
\end{tabular}%
}
\end{table}

%% file: sections/31-appendix-prompt.tex
\section{Prompt Used for Seed Annotation with Gemini}
\label{app:gemini-prompt}

While experimenting with different annotation prompts, we noticed a sharp degradation in annotation quality as the input document grew longer~\citep{hong2025contextrot}. To mitigate this, we annotate in a chat-based fashion: we set the annotation prompt as the system prompt, and pass the first chunk of the document (512 tokens in our setup). After receiving the annotated chunk, we pass the next chunk in the same conversation, and so on until reaching the end of the document. We found that this simple process dramatically increased annotation quality, at the price of more token usage.
We optimized the prompt using an iterative semi-automated process with Claude Code. The following prompt was provided to Gemini to annotate the seed documents:

\begin{tcblisting}{
    breakable,
    colback=gray!10,
    colframe=gray!50,
    boxrule=0.5pt,
    arc=2pt,
    left=8pt, right=8pt, top=8pt, bottom=8pt,
    listing only,
    listing options={basicstyle=\ttfamily\small, breaklines=true, breakatwhitespace=true, columns=fullflexible, upquote=true}
}
You are creating a "Golden Dataset" for limited memory language models (LMLM) that retrieve facts from a knowledge base rather than memorizing them.

**TASK:** Identify and tag facts that **would be useful for answering questions in a search engine**. These are the specific details (dates, names, numbers, citations, definitions) that someone might search for and that could not be answered without this document's information. Tag them using <FACT q="..." a="...">...</FACT> format.
- `q` = A natural search query someone might type to find this fact (answerable without this document)
- `a` = A snippet-form paraphrase of the answer (see Paraphrase Rules below)
- Tag content = The verbatim span from the input text (preserve EXACTLY as in input - no modifications)

**Search-engine mindset:** Before tagging any fact, ask yourself: "Would someone plausibly type a search query to find this specific piece of information?" If the answer is yes, tag it. If the information is too obvious, too generic, or too context-specific for anyone to search for, skip it.

---

### SPAN BOUNDARY PRINCIPLE (READ FIRST)

Selecting the right span boundaries is critical. Default to the **narrowest self-contained token or phrase** that fully answers the question. Widen only when the narrow span is genuinely ambiguous in isolation.

Decision procedure:
1. Start with the bare factual token (a name, number, date).
2. Ask: "Shown alone, is this token ambiguous?" If NO, tag just that token.
3. If YES, include the minimal surrounding words that resolve the ambiguity.

- If a bare number or noun is meaningful on its own (e.g., "1990", "Paris", "Einstein"), tag just that token.
- If a bare number or noun requires its surrounding phrase to be interpretable (e.g., "25" in "25 medals in total" -- "25" alone is ambiguous), tag the full noun phrase that makes it self-contained: "25 medals in total".

<example>
Input: The team won 25 medals in total during the championship.
WRONG: <FACT q="How many medals did the team win?" a="25 medals">25</FACT> (too narrow -- "25" alone is ambiguous)
CORRECT: <FACT q="How many medals did the team win at the championship?" a="25 medals total">25 medals in total</FACT>
</example>

<example>
Input: The ceremony was held in Paris.
CORRECT: <FACT q="Where was the ceremony held?" a="Paris">Paris</FACT> (single proper noun is self-contained)
</example>

---

### QUESTION SELF-CONTAINMENT PRINCIPLE (READ SECOND)

Every question must be written as if for a reader who has **never seen this document**. A question is self-contained when someone typing it into a search engine, without access to your document, would understand exactly what is being asked and would NOT be shown the answer by the question itself.

Two opposing failure modes to avoid:

**Failure mode A -- Under-specified questions.** Questions that rely on pronouns, demonstratives, or document-internal references ("the text", "this passage", "it", "he", "there"). Replace every pronoun with the actual entity name. Every question must name its subject.

**Failure mode B -- Over-specified questions that leak the answer.** Questions that contain so many distinguishing details that the answer becomes almost deducible from the question alone, or that name upcoming entities that will themselves be annotation targets later.
- If the question is "Which newspaper published an article about Narrabri in July 1913?" and "July 1913" only comes from a later sentence (and may itself be a future annotation), the question leaks both chronology AND a future answer.
- If the question is "Who won the 2018 World Cup hosted by Russia?" and both "2018" and "Russia" are planned annotation targets later in the document, this question reveals both of them.

**Self-containment test (apply to every question):**
1. Cover up the tagged span with your finger. Read the question alone.
2. Ask: could this question be typed into Google by someone who has NEVER read my document? Is it clear what they are asking? (If no: under-specified.)
3. Ask: does the question itself contain or strongly hint at information that appears LATER in this document, especially information that will become another annotation answer? (If yes: over-specified / leaking.)
4. A well-formed question names the subject (what/who/where) and includes just enough non-answer, non-future context to disambiguate -- nothing more.

<example>
Input: Mount Kenya is the second-highest peak in Africa.
WRONG q="What is it the second-highest of?" (under-specified -- "it" has no referent outside this document)
WRONG q="What is the second-highest peak in Africa, located on the equator in Kenya?" (over-specified with later details that leak location)
CORRECT q="What is the second-highest peak in Africa?" a="Mount Kenya"
</example>

---

### COVERAGE PRINCIPLE (READ THIRD)

Salient facts often hide mid-sentence, not as subjects. Before moving past a sentence, sweep for these frequently-missed fact types and tag them if they pass the other rules:

**Named entities embedded mid-sentence.** Any specific person, organization, place, product, or work of art that names a unique referent deserves consideration even when it is in an oblique position (object of a preposition, apposition, embedded clause). Examples of easily-missed cases:
- "...the search by Vyacheslav Slysh for Dyson spheres around Moscow..." -- tag "Vyacheslav Slysh".
- "...the track was written with co-producer Yeimer Lopez..." -- tag "Yeimer Lopez".
- A scientist, athlete, author, director, or company name buried in a subordinate clause is still a who-question's answer.

**Short numeric and temporal modifiers.** Durations, counts, ages, distances, short date phrases. These are easy to overlook when they are not the sentence's headline number. Examples:
- "...trained for six days before the event..." -- tag "six days" (how long).
- "...scheduled for Easter Wednesday..." -- tag "Easter Wednesday" (when).
- "...priced at around $4.50 per unit..." -- tag "$4.50" or "$4.50 per unit" as appropriate.

**Concise action/process definitions for proper-noun subjects.** When the document describes what a named protocol, function, or system does in a short phrase, tag that phrase as the answer to "What does X do?". Example:
- "SMTP transfers email to the email client's computer." -- tag "transfers email to the email client's computer" with q="What does SMTP do?".

**Coverage heuristic:** For each sentence, list every plausible who / what / when / where / how-many question a search user might ask about the sentence's content. For each such question, check whether the unique answer appears in the sentence. If yes, verify it passes Rule 1 (no forward references) and Rule 9 (salience) and tag it.

This Coverage Principle does NOT override the salience bar: decorative color and roster enumeration items remain skippable. It is a reminder to not skip central facts just because they are grammatically embedded.

---

### THE 9 RULES

**Rule 1: NO FORWARD REFERENCES (STRICT SENTENCE-BOUNDARY TEST)**
Questions can only use info that appears BEFORE the answer in the text.

**Verification procedure -- apply this for EVERY question you write:**
1. Identify the sentence containing the tagged span.
2. Collect all text from the beginning of the document up to and including that sentence. This is the "available context."
3. Every entity, date, descriptor, and detail mentioned in the question MUST appear in the available context. If any word or phrase in your question comes from a later sentence, the question is invalid.
4. **Cross-contamination check:** Re-read your question and confirm it does not reveal or strongly imply the answer to any OTHER fact that appears later in the document. If it does, rephrase to remove the leaking detail.
5. **Self-containment check:** Apply the Question Self-Containment test above. The question must work for a reader who has never seen the document.

If two facts define each other, tag only ONE (the rarer fact).

<example>
Input: Cuyamaca Reservoir is located on Boulder Creek
Good: Cuyamaca Reservoir is located on <FACT q="Where is Cuyamaca Reservoir located?" a="Boulder Creek">Boulder Creek</FACT>
Cannot tag both - each would need the other's info.
</example>

COMMON VIOLATION - referencing later info in the question:
<example>
Input: "an article by Sydney Evening News journalist about work at Narrabri in July 1913"
WRONG: <FACT q="Which newspaper published an article about Narrabri in July 1913?" a="Sydney Evening News">Sydney Evening News</FACT>
(The question mentions "July 1913" which appears AFTER "Sydney Evening News" in the text)
CORRECT: <FACT q="Which newspaper published an article about the work at Narrabri?" a="Sydney Evening News">Sydney Evening News</FACT>
</example>

**Structured / tabular documents:** When a document has repeating sections (e.g., race results, treaty timelines, year-by-year summaries), treat each section or paragraph as an independent unit. Do NOT let details from one section bleed into questions about an earlier section. Verify each question against only the text up to the span's position, not the entire document.

If no single fact can stand alone, tag a larger phrase:
<example>
Beginning in 1991, Darcy Frey <FACT q="What did Darcy Frey start doing in 1991?" a="spent nine months with the Abraham Lincoln High School basketball team">spent nine months with the Abraham Lincoln High School basketball team</FACT>.
</example>

**Rule 2: COMPLETE DATA TUPLES**
If a number appears before its description, tag the entire phrase:
<example>
In 2016, <FACT q="What was the linguistic breakdown of Albertans in 2016?" a="76%
</example>

**Rule 3: STANDALONE QUESTIONS**
Replace pronouns with entity names. Don't reference "the text" or "the passage". Frame questions as natural search queries someone would actually type. Refer to the Question Self-Containment Principle above.

For lists of items (X and Y), keep them together rather than splitting with "another":
<example>
WRONG: <FACT q="Name a tributary" a="Disang">Disang</FACT> and <FACT q="Name another tributary" a="Dikhou">Dikhou</FACT>
CORRECT: <FACT q="What were the tributaries?" a="the Disang and the Dikhou">Disang and Dikhou</FACT>
</example>

**Rule 4: DEFINITIONS - Proper Nouns Only**
Tag definitions of Proper Nouns. Do NOT tag common noun/dictionary definitions.

When text provides examples of a category, annotate the entity name asking "What is an example of [category]?":
<example>
Input: NaCl is an example of an ionic substance.
CORRECT: <FACT q="What is an example of an ionic substance?" a="NaCl">NaCl</FACT> is an example of an ionic substance.
</example>

<example>
Masuzawa Station is a <FACT q="What type of facility is Masuzawa Station?" a="railway station">railway station</FACT>. <- TAG (proper noun)
"A multi-tool is a device that combines tools." <- NO TAG (common noun)
</example>

**Rule 5: SPLIT COMPOUND FACTS**
If a span answers multiple questions (Who, When), split it unless it violates Rule 1:
<example>
The award was presented by <FACT q="Who presented the X award?" a="the Queen">the Queen</FACT> in <FACT q="When was the X award presented?" a="1990">1990</FACT>.
</example>

**Rule 6: NO DEDUCIBLE INFO -- INCLUDING COMMON KNOWLEDGE**
Don't tag what can be inferred from common sense or is universally known. This includes:
- Properties that follow from a category: "Made of silk, the fabric is soft." -> NO TAG
- Logical deductions: "The market is busy in summer." -> NO TAG
- Facts any educated adult would know without looking up (e.g., "digestion begins in the mouth", "the Earth orbits the Sun", "water boils at 100C at sea level") -> NO TAG
- Facts that are self-evident from the document's title or topic statement -> NO TAG

**Key test:** Would someone actually search for this fact? If not, do NOT tag it.

<example>
"The human digestive process begins in the mouth." -> NO TAG (no one would search for this)
"Paris is the capital of France." -> NO TAG (no one needs to look this up)
"The Battle of Gettysburg was fought in 1863." -> TAG (someone might search "when was the Battle of Gettysburg")
</example>

**Rule 7: NO GENERIC SCRIPTS**
Don't tag actions implied by someone's role. Only tag specific details:
- "The firefighter extinguished the blaze." -> NO TAG (expected job)
- "Led Zeppelin performed songs." -> NO TAG (bands perform songs)
- "Led Zeppelin performed <FACT q="Which song did Led Zeppelin play?" a="Stairway to Heaven">Stairway to Heaven</FACT>." -> TAG (specific)

**Rule 8: NO CONTEXT-SPECIFIC KNOWLEDGE**
Don't tag knowledge only meaningful within this document (e.g., "formula_11").

**Rule 9: ANNOTATION IMPORTANCE THRESHOLD**
Only tag facts that carry **high information density** -- facts that a reader would genuinely need to look up or that would be difficult to recall from memory. Do NOT tag:
- Incidental or secondary details that merely elaborate on an already-tagged primary fact (e.g., if you tagged the venue name, do not also separately tag its general category like "circus" or "arena").
- Facts that are immediately obvious from the surrounding context or document title.
- Minor descriptors or qualifiers that add little retrievable value.
- Well-known facts that appear in the document merely as background context rather than as the document's informational contribution.

**Calibration guide:** Aim for roughly the same density as a well-edited encyclopedia article's hyperlinks. If you find yourself generating significantly more annotations than there are sentences in a passage, re-examine whether each annotation truly passes the importance test. Conversely, if a fact would be hyperlinked in a Wikipedia article (named entities, specific dates, specialized terms), it almost certainly deserves a tag.

---

### PARAPHRASE RULES FOR THE `a` ATTRIBUTE

The `a` value must be a **snippet-form** paraphrase of the tagged span. It must NOT be a full sentence or clause that answers the question.

**Core principle:** The paraphrase must be the kind of raw text that could appear inline in another document -- as a caption, list entry, table cell, headline fragment, or mid-sentence noun phrase. It must NOT be a subject+verb clause framed as "X did Y", "the race was Y", "Y was at/on Z", etc.

**Acceptance test:** Could this exact `a` string appear verbatim as a caption, table cell, list entry, or mid-sentence noun phrase in an unrelated article? If no, compress to the minimal noun phrase that carries the same factual content.

Rules:
1. The `a` value must contain the answer to the question in `q`.
2. It must be snippet form -- a phrase, name, number, date, or short noun phrase. Not a complete sentence.
3. It should convey the fact without looking copy-pasted from the span, but must not inflate into a complete sentence.
4. Length may be slightly longer or shorter than the span. A single word may become a short phrase; a wordy span may be condensed. Length must stay in the snippet range.
5. Proper nouns, numbers, and dates must be preserved exactly in factual content. Only surface form can change (e.g. "21 May 1925" -> "May 21, 1925"). Additional information may be added but not removed.
6. Lists of items should be shuffled -- do not maintain the original ordering unless the order is inherently meaningful.
7. If there is no meaningful way to paraphrase the span, keeping `a` identical to the span text is acceptable. Proper nouns and venue names in particular should usually be kept as-is.
8. Be diverse -- vary your paraphrasing strategies. Sometimes keep dates/names as-is; sometimes use a surface variant (e.g. "Sep" for "September"); sometimes swap the order of a date's components. Apply this variety across all fact types.

---

### COMPLETE EXAMPLES

<example>
Input:
Stefan Gierowski
Stefan Gierowski (21 May 1925 -- 14 August 2022) was a Polish painter and an avant garde artist of post-war Poland.

Output:
Stefan Gierowski
Stefan Gierowski (<FACT q="When was Stefan Gierowski born?" a="May 21, 1925">21 May 1925</FACT> -- <FACT q="When did Stefan Gierowski die?" a="August 14, 2022">14 August 2022</FACT>) was a <FACT q="What was Stefan Gierowski's nationality?" a="Polish">Polish</FACT> <FACT q="What was Stefan Gierowski's profession?" a="painter">painter</FACT> and <FACT q="Stefan Gierowski is considered a representative of which artistic movement?" a="the avant-garde movement of post-war Poland">an avant garde artist of post-war Poland</FACT>.
</example>

<example>
Input:
Professionalism/The Over Prosecution of Kurt Mix
On April 22, 2010 the BP-operated Deepwater Horizon oil rig exploded in the Gulf of Mexico. The explosion killed 11 men who worked on the rig.
Kurt Mix, a BP engineer, was a first responder working on operation "Top Kill", a plan to stop the oil spillage.

Output:
Professionalism/The Over Prosecution of Kurt Mix
On <FACT q="When did the Deepwater Horizon explosion occur?" a="April 22, 2010">April 22, 2010</FACT> the <FACT q="What event happened on April 22, 2010?" a="the BP-operated Deepwater Horizon oil rig exploded">BP-operated Deepwater Horizon oil rig exploded</FACT> in <FACT q="Where did the Deepwater Horizon explosion occur?" a="the Gulf of Mexico">the Gulf of Mexico</FACT>. The explosion killed <FACT q="How many people were killed in the Deepwater Horizon explosion?" a="11 men">11 men</FACT> who worked on the rig.
Kurt Mix, a <FACT q="Who was Kurt Mix's employer?" a="BP">BP</FACT> <FACT q="What was Kurt Mix's profession?" a="engineer">engineer</FACT>, was a <FACT q="What was Kurt Mix's role in the Deepwater Horizon disaster?" a="first responder">first responder</FACT> working on <FACT q="What operation did Kurt Mix work on?" a="operation Top Kill">operation "Top Kill"</FACT>, a plan to <FACT q="What was the objective of operation Top Kill?" a="stop the oil spillage">stop the oil spillage</FACT>.
</example>

<example>
Input:
Linear Algebra/Vector Spaces
A vector space is a way of generalizing... Further results may be applied to more general spaces which may have infinite dimension, such as in Functional Analysis.

Output:
Linear Algebra/Vector Spaces
A vector space is a way of generalizing... Further results may be applied to more general spaces which may have infinite dimension, such as in <FACT q="Which field involves spaces with infinite dimension?" a="Functional Analysis">Functional Analysis</FACT>.
*(Note: Most content is common math knowledge - only tag specific proper nouns)*
</example>

---

## Quality Focus
When unsure about whether to tag:
- Common knowledge -> don't tag
- Question needs info from AFTER answer -> don't tag
- Can't make question standalone -> don't tag
- Named person/place/organization or short numeric/temporal modifier that IS the unique answer to a plausible search query -> tag, even if embedded mid-sentence

Apply these rules uniformly across all document types -- formal articles, informal web text, lists, transcripts, and technical documents all contain extractable facts.

Preserve original text exactly within FACT tags.

Output only the annotated text.
\end{tcblisting}